\documentclass[letterpaper]{article} 
\usepackage[preprint]{aaai2027}  
\usepackage[hyphens]{url}  
\usepackage{graphicx} 
\usepackage{natbib}  
\usepackage{caption} 
\usepackage{algorithm}
\usepackage{algorithmic}
\usepackage{amsmath}
\usepackage{xcolor} 
\usepackage{amssymb}
\usepackage[table]{xcolor}
\definecolor{firstcolor}{HTML}{B7A4D8}
\definecolor{secondcolor}{HTML}{D8D5E6}
\definecolor{thirdcolor}{HTML}{F1F0F6}

\usepackage{newfloat}
\usepackage{listings}
\DeclareCaptionStyle{ruled}{labelfont=normalfont,labelsep=colon,strut=off} 
\floatstyle{ruled}
\newfloat{listing}{tb}{lst}{}
\floatname{listing}{Listing}

\usepackage{booktabs}

\title{JewelTry: Mask-Free Scale Aware Jewelry Virtual
Try-On}
\author{
    Xinlei Niu\textsuperscript{\rm 1}\thanks{Work done during the internship at Amazon.},
    Peixia Li\textsuperscript{\rm 2},
    Jun Wang\textsuperscript{\rm 2},
    Chenchen Xu\textsuperscript{\rm 2},
    Jiayu Yang\textsuperscript{\rm 2},
    Jing Zhang\textsuperscript{\rm 1},
    Pulak Purkait\textsuperscript{\rm 2},
    Hongdong Li\textsuperscript{\rm 1,\rm 2}
}
\affiliations{
    \textsuperscript{\rm 1}Australian National University, Canberra, Australia;\\
    \textsuperscript{\rm 2}Amazon, Melbourne, Australia\\

    xinlei.niu@anu.edu.au
}

\begin{document}

\maketitle

\begin{abstract}
Virtual try-on (VTON) enables customers to visualize how fashion products appear when worn and has become an important technology for online shopping. While recent advances have substantially improved garment VTON, jewelry remains a challenging and underexplored category due to its small size, rigid structure, and sensitivity to fine-grained visual details. Realistic jewelry VTON requires not only faithful appearance transfer but also accurate scale and placement relative to the wearer. Existing jewelry VTON methods typically rely on mask guidance, whereas mask-free approaches lack explicit guidance for modeling the product scale. To bridge this gap, we introduce \textbf{JVTO-Bench}, a benchmark dataset for scale-faithful jewelry VTON, providing reference–source–target triplets with real-world product-scale annotations across four major jewelry categories. Building upon this benchmark, we propose \textbf{JewelTry}, a mask-free diffusion framework for scale-aware jewelry VTON. JewelTry incorporates a scale adapter that encodes product dimensions into a scale token, enabling the model to learn scale relationships between jewelry items and surrounding human anatomy in-context. To further improve jewelry consistency, we introduce a single-directional condition attention mechanism and an attention refinement loss that preserve both coarse geometry and fine-grained structural details of the reference jewelry. Extensive experiments show that JewelTry achieves a balance among visual fidelity, background preservation, object consistency and scale accuracy, establishing a strong baseline for mask-free, scale-aware jewelry virtual try-on.

\end{abstract}


\section{Introduction}

Visual understanding plays a critical role in online fashion shopping, where customers rely heavily on product image to evaluate appearance, fit, and purchasing suitability. Virtual try-on (VTON) have emerged as an effective solution for bridging the gap between product presentation and real-world appearance, enabling customers to visualize how fashion items look when worn. While substantial progress has been made in garment VTON, jewelry remains a considerably more challenging category due to its small size, rigid structure, and sensitivity to fine-grained visual details~\cite{miao2025shining}. Beyond its design, the appeal of jewelry depends on subtle factors such as its scale, placement, and harmony with the wearer. Without realistic try-on imagery, customers must infer these factors from standalone product photos and textual metadata such as dimensions and materials, which can lead to misleading expectations and reduced purchase confidence. Therefore, scale-faithful jewelry VTON is a practically important and technically challenging problem: it requires preserving the detailed appearance of small objects while rendering them at physically plausible sizes and locations on diverse human models.

\begin{figure}[t]
    \centering
    \includegraphics[width=\linewidth]{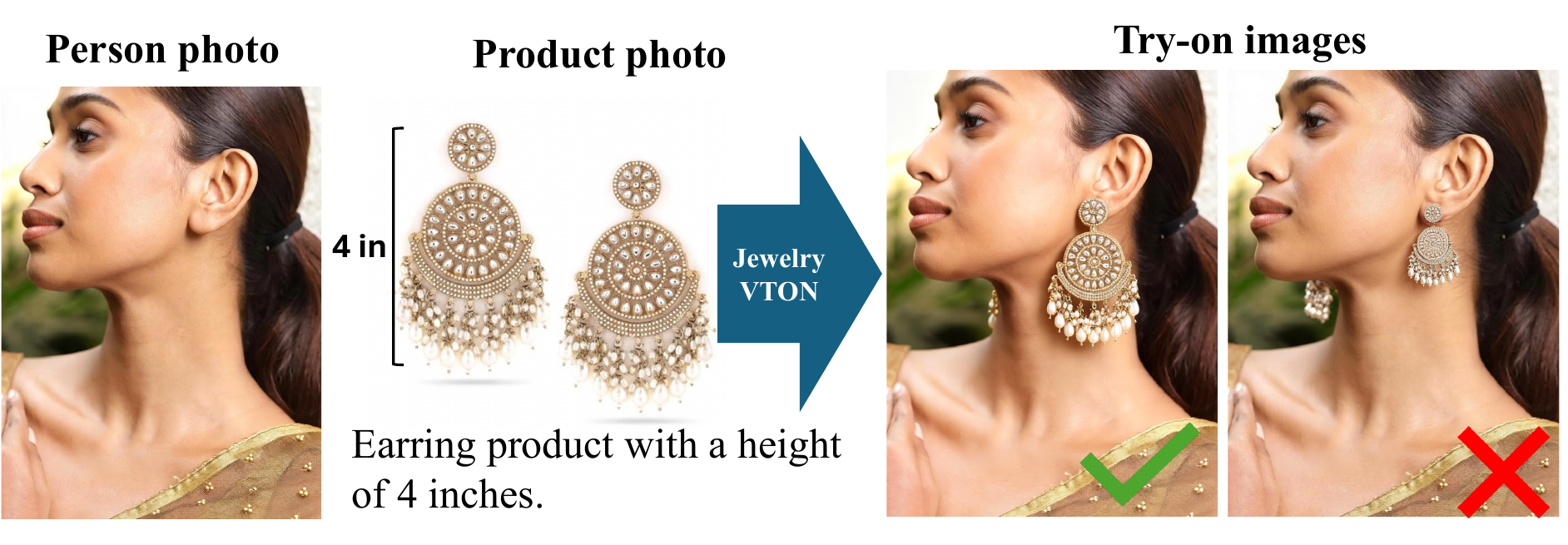}
    \vskip -0.1in
    \caption{Mask-free scale aware jewelry VTON provides more realistic visualization of product size and appearance.}
    \label{fig:sur}
    \vskip -0.2in
\end{figure}

Recent advances in image generation and editing have substantially improved virtual try-on for garments, accessories, and jewelries~\cite{wu2025qwen,li2025refVTON,zhu2023tryondiffusion,xu2025ootdiffusion,choi2024improving,feng2025omnitry,miao2025shining}. In garment VTON, recent methods have moved toward parser-free or mask-free approaches, reducing the need for dense human parsing or manually specified spatial guidance~\cite{zhang2025boow,feng2025omnitry,du2025all}. However, mask-free jewelry VTON remains comparatively underexplored. This setting is especially challenging, as the small, rigid, and detailed nature of jewelry demands accurate scale, precise placement, and faithful geometry preservation. Errors in size, shape, or location can make a try-on image visually misleading to customers. Existing jewelry VTON methods typically rely on mask guidance to constrain jewelry position and scale~\cite{miao2025shining}, while more general try-on frameworks do not explicitly model real-world product scale and instead require the model to infer it implicitly from visual appearance~\cite{feng2025omnitry}. These limitations motivate the need for a mask-free framework that can preserve product appearance, placement, and physical scale without relying on manually specified masks or pixel-level spatial guidance.

A major obstacle to advancing jewelry VTON is the lack of suitable benchmarks. Existing VTON datasets primarily target garments or general accessories, and they rarely provide jewelry-specific annotations or reliable real-world product-scale information~\cite{hu2026garments2look}. As illustrated in Figure~\ref{fig:sur}, scale is particularly critical in jewelry VTON: the perceived realism of a jewelry depends not only on transferring its visual appearance, but also on rendering it at a plausible size relative to the wearer's anatomy. To bridge this gap, we introduce JVTO-Bench, a benchmark dataset designed specifically for scale-faithful jewelry VTON. JVTO-Bench covers four major jewelry categories and represents each sample to a reference-source-target triplet with accurate real-world product scale. JVTO-Bench provides both training and test splits, enabling model development as well as evaluation. By pairing triplet-based try-on supervision with product-scale information, JVTO-Bench supports training and assessment of jewelry VTON models under mask-free conditions.

Building on JVTO-Bench, we propose JewelTry, a mask-free and scale-aware framework for jewelry VTON. JewelTry is designed to address two fundamental challenges in jewelry VTON: (1) rendering jewelry at a physically plausible scale and (2) preserving the fine-grained details of the jewelry. To model real-world scale without relying on manually specified masks or explicit geometric supervision, we introduce a scale adapter that encodes product dimensions, measured in inches, into a scale token. The scale token enables the model to learn relative scale relationships between jewelry items and surrounding anatomical regions, such as ears, fingers, wrists, and necks, resulting in more realistic and scale-consistent try-on results. To improve jewelry fidelity, we further introduce a single-directional condition attention mechanism and an attention refinement loss. The single-directional condition attention prevents reference condition tokens from being contaminated by noisy latent features during the diffusion process, thereby preserving the coarse structure of the reference jewelry. Building upon this, the attention refinement loss explicitly supervises the interaction between jewelry condition tokens and the target try-on region, encouraging the model to focus on relevant object areas and improving fine-grained appearance consistency. Together, these components enable JewelTry to faithfully preserve both the physical scale and visual structure of jewelry in a fully mask-free setting.
In summary, our main contributions are threefold:
\begin{itemize}
    \item \textbf{JVTO-Bench Dataset.} A benchmark dataset for the mask-free jewelry VTON task, featuring high-quality reference-source-target image triplets as well as product-scale annotations across four major jewelry categories.
    \item \textbf{JewelTry.} A diffusion framework designed for jewelry consistency and scale awareness in a mask-free manner.
    \item We conduct extensive experiments demonstrating that JewelTry establishes a state-of-the-art baseline for scale-aware and mask-free jewelry VTON.
\end{itemize}

\begin{figure*}[t]
    \centering
    \vskip -0.1in
    \includegraphics[width= 0.95\textwidth]{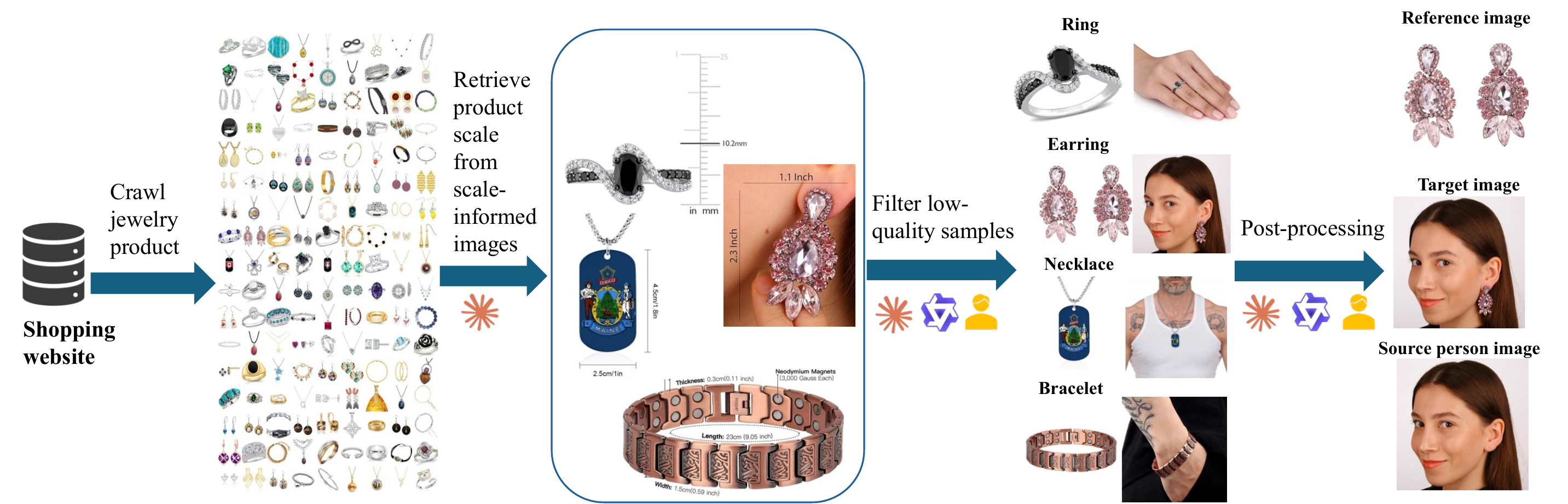}
    \vskip -0.05in
    \caption{Overview of the JVTO-Bench dataset preparation pipeline.}
    \label{fig:dataprocess}
    \vskip -0.15in
\end{figure*}

\section{Related Work}

\textbf{Object-guided Image Editing and Generation.}
Object-guided image editing and generation aims to modify or synthesize an image according to a given reference object, while preserving scene context and visual realism. The generated object is expected to match the reference in appearance, structure, and semantics, and be naturally integrated into the target image~\cite{tan2025ominicontrol,zhang2025easycontrol,chen2026attrictrl,chen2025xverse,xu2026towards,she2025mosaic,cheng2025umo,zhang2026your,shin2025exploring}. Virtual try-on is a specialized task of object-guided editing, where the reference object is a wearable item such as clothing and accessories, and the target image is a person image. Compared with general object-guided editing, VTON imposes stronger constraints on geometric alignment, scale consistency, and interaction with human body regions, since the generated object must be both realistic and correctly positioned with proper proportions.

\textbf{Garment Virtual Try-On.}
Garment VTON aims to generate a realistic image of a person wearing a target garment by transferring clothing appearance from a reference image to the person image. It has evolved from early U-Net-based reconstruction methods~\cite{issenhuth2019end} to diffusion-based frameworks with substantially improved generation quality and garment realism. Recent methods enhance garment authenticity through specialized adapters and semantic feature injection~\cite{wang2024fldm,choi2024improving}, while diffusion transformer-based approaches further improve scalability and the modeling of complex physical deformations~\cite{lee2025voost,meiphysdiff} with inpainting-based masking strategies for paired data generation~\cite{jiang2024groupdiff}. More recently, garment VTON has shifted from parser-based pipelines to mask-free paradigms to enable realistic try-on without explicit segmentation masks~\cite{niu2024pfdm,du2025all,zhang2025boow,dumitigating,li2025refVTON,Kwon_2026_CVPR}. Beyond garments, omni-style VTON further extends virtual try-on to accessories and diverse object categories through unified mask-free frameworks~\cite{feng2025omnitry,wang2025jco,zeng2026omnidit}.

\textbf{Jewelry and Ornament Virtual Try-On.}
Despite the progress in garment VTON, high-end jewelry remain challenging due to their rigid shapes, details, and complex topologies~\cite{chang2024glamtry}. ShiningYourself~\cite{miao2025shining} pioneers mask-guided ornament VTON, where a target mask is required to explicitly specify the placement region and object scale during generation. SparklingTogether~\cite{xu2026sparkling} further extends the mask-guided single jewelry VTON to a mask-guided multi-accessory VTON framework. Recent omni-style methods~\cite{feng2025omnitry,wu2025qwen} support mask-free jewelry VTON, but they are not specifically designed for enhancing jewelry scale-faithful, where accurate scale control and structural consistency are critical.

\section{JVTO-Bench Dataset}

Accurate product-scale information is essential for the mask-free jewelry virtual try-on task, as it directly affects size realism and visual plausibility. However, existing jewelry VTON research lacks publicly available datasets with reliable scale annotations. To address this gap, we introduce \textbf{JVTO-Bench}, a benchmark dataset for scale-faithful jewelry VTON. JVTO-Bench covers four major categories: rings, earrings, necklaces, and bracelets, and contains over 23K high quality and diverse jewelry fashion images, ranging from well-posed shop images to unconstrained consumer photos. Each sample is organized as a triplet ${J,P,T}$ with product scale in inches, where $J$ is the reference jewelry image, $P$ is the source try-off image, and $T$ is the target try-on image. We position JVTO-Bench as a key contribution of this work and a comprehensive resource for future jewelry VTON research.
As illustrated in Figure~\ref{fig:dataprocess}, our dataset construction pipeline consists of four stages:

\textbf{Stage 1: Data Collection.}
To construct a diverse dataset, we collect jewelry data from a representative online shopping platform across major markets, including the UK, the US, and India, ensuring broad coverage of jewelry styles and regional preferences. Each item contains one main display image and several auxiliary images. These auxiliary images may include valid human try-on images as well as product-only, detail, lifestyle, or scale-informed images, yielding approximately 3M items in the raw pool.

\textbf{Stage 2: Product Scale Retrieval.}
Although sellers often provide product scale information, it is often noisy or inaccurate. We use Claude-Sonnet-4 to extract scale annotation directly from scale-informed images followed by human validation. See our supplementary material for more details.

\textbf{Stage 3: Data Cleaning.}
We apply strict filtering to obtain high-quality product-person pairs. For each reference image, we require exactly one jewelry product on a clean white background. For each try-on image, the jewelry must be clearly visible and correctly matched to the reference. We use Claude-Sonnet-4 and Qwen-VL-3 for automatic filtering, followed by expert-level manual verification to further clean the mismatched samples and ensure quality.

\textbf{Stage 4: Annotation and Post-processing.}
We convert filtered product-person pairs into training-ready triplets. The reference jewelry image is tightly cropped around the product region with an additional margin $25\%$ using Grounding DINO~\cite{liu2024grounding}. We also use Claude-Sonnet-4 to generate captions for try-on images, following the template: ``The person from image 1 is wearing/holding \{jewelry type\} from image 2 \{optional position descriptor\}.'' These captions describe the interaction between the person and jewelry and can serve as text prompts for training.
Since source images without the jewelry are typically unavailable, we synthesize try-off source images by removing the jewelry. Existing segmentation-and-inpainting pipelines~\cite{feng2025omnitry} are less effective for jewelry removal, as residual shadows often remain, causing visible artifacts and degraded realism. We further refine the try-off images using the output of Qwen-Object-Remover\footnote{https://huggingface.co/prithivMLmods/Qwen-Image-Edit-2511-Object-Remover} and the SSIM-based difference map between the object-removed result and the original target image, which helps localize residual jewelry and shadow artifacts for inpainting. See our supplement for more details.

\section{Method}
\begin{figure*}[t]
    \vskip -0.1in
    \centering
    \includegraphics[width=\textwidth]{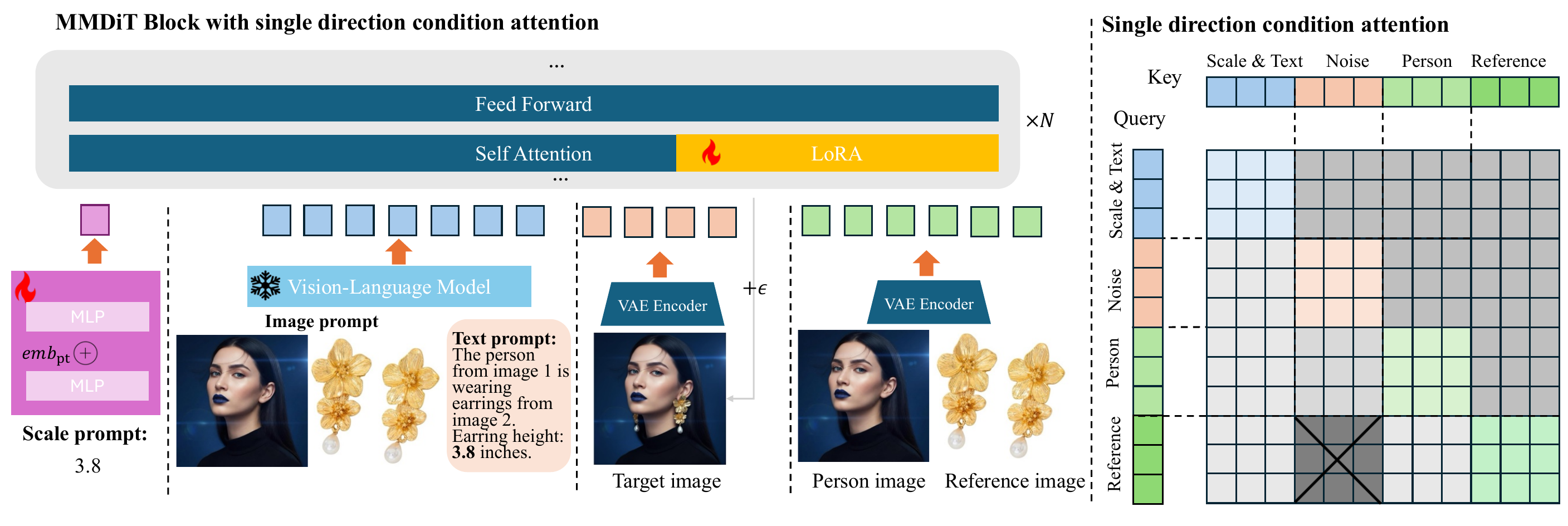}
    \vskip -0.15in
    \caption{Overview of the JewelTry framework. Snowflake and fire icons denote frozen and trainable parameters, respectively.}
    \label{fig:overview}
    \vskip -0.15in
\end{figure*}

We now present the technical details of our method, JewelTry. Figure~\ref{fig:overview} provides the framework overview, which consists of three key components: (1) a scale adapter that projects numerical scale prompts into scale tokens; (2) a single-directional condition attention mechanism that prevents reference feature collapse and preserves jewelry structure consistency; and (3) an attention refinement loss that explicitly encourages the model to focus on fine-grained details during training.

\subsection{Scale Adapter}\label{sec:scale_adapter}
The first challenge is enabling the model to perceive product scale, which is essential for generating realistic try-on results with accurate size and proportion. As in Figure~\ref{fig:overview} (right), we first incorporate scale information into the text prompt of a vision-language model (VLM), leveraging its reasoning capability to interpret physical measurements and translate them into semantic scale-aware guidance. To further make the model aware of scale inputs, we involve a scale adapter that encodes product’s physical measurement in inches into a scale token~\cite{chen2026attrictrl,xu2025numerikontrol}. The scale token is subsequently concatenated with the text token produced by the VLM.
The scale adapter is motivated by the need to model the contextual relationship between real-world jewelry dimensions and human anatomy. Since reference product images and target person images are typically captured under different camera settings and viewpoints, recovering exact camera parameters or performing explicit geometric alignment is non-trivial. Instead of enforcing direct absolute scale supervision, the scale adapter provides an implicit condition that encourages the model to learn relative scale relationships between jewelry size and nearby anatomical regions. 

\subsection{Single Directional Condition Attention}\label{sec:single_attn}
\begin{figure}[t]
    \centering
    \includegraphics[width=\linewidth]{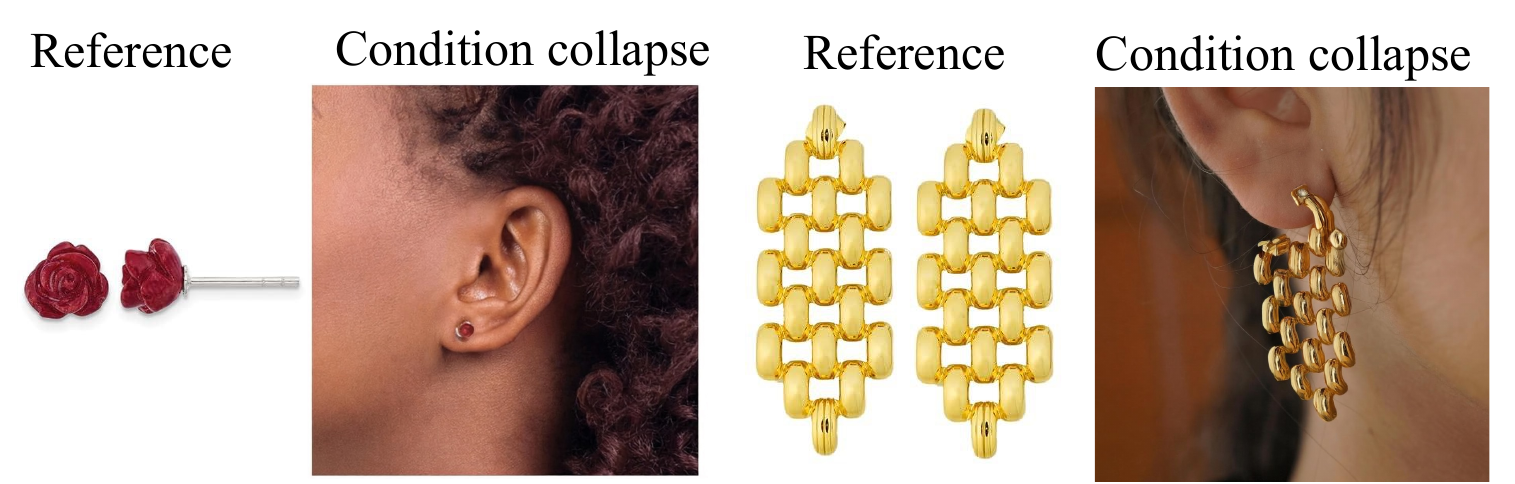}
    \vskip -0.05in
    \caption{Examples of condition collapse problem, where the results fail to preserve the rigid structure in reference images.}
    \label{fig:condition_collapse}
    \vskip -0.2in
\end{figure}

Jewelry items typically have rigid structures and fixed topology, making structural consistency a critical challenge in VTON~\cite{miao2025shining}. The generated result must faithfully preserve the jewelry structure and fine-grained details from the reference image. In bidirectional self-attention, noisy latent tokens and condition tokens attend to each other symmetrically. While this facilitates information exchange, it can also destabilize the conditioning representation: fine-grained jewelry cues encoded from the reference image may be corrupted by noisy latent features. We refer to this as \emph{condition collapse} (Figure~\ref{fig:condition_collapse}), where the model fails to consistently preserve the structural details of the reference jewelry.

To mitigate condition collapse, we introduce single direction condition attention in MMDiT to preserve a stable jewelry conditioning. Standard bidirectional self-attention allows noisy latent tokens and condition tokens to update each other symmetrically. While effective for general information exchange, this design is suboptimal for jewelry VTON: the reference jewelry tokens encode rigid topology and fine-grained product details, and should remain a reliable source of structural guidance rather than being updated by noisy latent features. We therefore block the reverse attention path from noisy latent tokens to jewelry tokens, while preserving the forward guidance from jewelry tokens to the noisy latent representation. In this way, latent tokens can still attend to the reference jewelry and receive structural guidance, whereas jewelry tokens are protected from noise-dependent interference. This asymmetric design maintains the stability of the jewelry representation during training, reducing condition collapse issue and improving structural fidelity. We provide additional discussion in our supplementary material.

Let $M$ denote the binary attention mask that controls the allowable interactions among condition and latent branches. We concatenate the query, key, and value tokens as
\begin{equation}
Q = [Q_{\text{s,t}}; Q_{\text{noise}}; Q_{\text{P}}; Q_{\text{J}}], \quad
K = [K_{\text{s,t}}; K_{\text{noise}}; K_{\text{P}}; K_{\text{J}}],
\nonumber
\end{equation}
\begin{equation}
V = [V_{\text{s,t}}; V_{\text{noise}}; V_{\text{P}}; V_{\text{J}}],
\nonumber
\end{equation}
where $\text{s,t}$ denotes scale and text tokens, $\text{P}$ denotes person-image tokens, and $\text{J}$ denotes jewelry-image tokens. For token positions $i$ and $j$, the masked attention score is computed as
$
A_{ij} = \frac{Q_i K_j^\top}{\sqrt{d_k}} + M_{ij}.
$
As illustrated in Figure~\ref{fig:overview} (right), our single direction condition attention mask is defined as
\begin{equation}
M_{ij} =
\begin{cases}
-\infty, & i \in \text{J},; j \in \text{noise}, \\
0, & \text{otherwise}.
\end{cases}
\end{equation}
Although attention masking in MMDiT has been explored for conditional generation and editing~\cite{cai2025ditctrl,wang2025jco,shen2025qk,zhang2025easycontrol,chen2026attrictrl}, existing methods mainly use it for spatial control, condition injection, or controllable generation. In contrast, our masking strategy targets condition collapse in jewelry VTON. Specifically, it prevents noisy latent tokens from interfering with fine-grained jewelry condition tokens while preserving the guidance from jewelry tokens to the latent branch, thereby improving jewelry structural consistency.

\subsection{Training Loss}
\begin{figure}
    \centering
    \includegraphics[width=\linewidth]{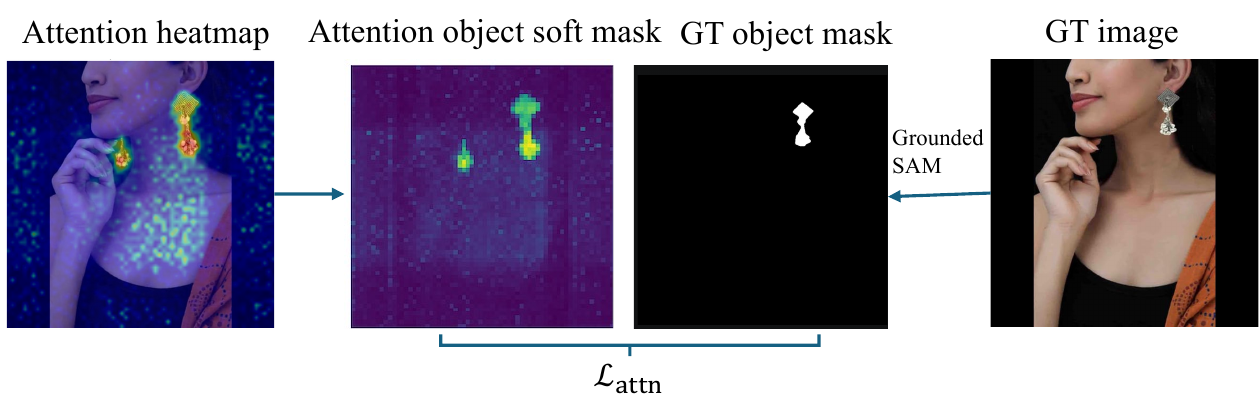}
    \vskip -0.1in
    \caption{Attention mask from later MMDiT blocks are supervised by the GT mask to preserve fine-grained detail.}
    \label{fig:attn_loss}
    \vskip -0.1in
\end{figure}
\textbf{Attention refinement loss.} 
To further improve the structural and fine-grained jewelry consistency in the try-on results, we introduce an attention refinement loss. We extract a predicted jewelry soft mask from the attention map between the reference query \(Q_{\text{ref}}\) and the noisy latent key \(K_{\text{noise}}\), and supervise it with the ground-truth jewelry mask. We observe that, in the later MMDiT blocks, this attention map naturally highlights the target jewelry region and is strongly correlated with the final generated result (see supplementary for more details). By explicitly aligning this attention-derived soft mask with the ground-truth object region, the proposed loss encourages more accurate spatial correspondence between the reference jewelry and the generated try-on image, leading to improved object placement, scale fidelity, and fine-grained structural preservation. As illustrated in Figure~\ref{fig:attn_loss}, the attention refinement loss $\mathcal{L}_{\text{attn}}$ is defined as
\begin{equation}
    \mathcal{L}_{\text{attn}}=  \mathcal{L}_{\text{BCE}}(A_i^{(k)},G)+  \mathcal{L}_{\text{Dice}}(A_i^{(k)},G)
\end{equation}
Where $A_i^{(k)}$ denotes the soft attention mask extracted from the $k$-th attention block at the denoising step $i$. $A_i^{(k)}$ is computed by averaging the cross-attention weights across all attention heads, aggregating them over the conditioning token dimension, and finally applying min–max normalization to obtain a continuous mask with values in $[0,1]$. $G$ denotes the ground-truth mask.
\begin{equation}
    \mathcal{L}_{\text{BCE}}(A,G) = - \frac{1}{N} \sum^N_{p=1} [G_p log A_p +(1-G_p)log(1-A_p)] \nonumber
\end{equation}
\begin{equation}
    \mathcal{L}_{\text{Dice}}(A,G) = 1- \frac{2\sum_{p=1}^N A_p G_p + c}{\sum_{p=1}^N A_p +\sum_{p=1}^N G_p + c} \nonumber
\end{equation}
$N$ represents the number of pixel and $c$ is a small constant.

\textbf{Object consistency loss.}
We also adopt the velocity prediction loss~\cite{lipman2023flowmatchinggenerativemodeling} and introduce an object-region loss that emphasizes the prediction accuracy within the target jewelry region:
\begin{equation}
    \mathcal{L}_{\text{obj}}
    =
    \mathbb{E}_{x_0 \sim \mathcal{D},\, x_1,\, t}
    \left[
    \left\|
    G \odot \left(
    v_\theta(x_t, t, c) - v_t
    \right)
    \right\|_2^2
    \right],
\end{equation}
where \(G\) denotes the ground-truth jewelry mask, \(v_\theta(x_t,t,c)\) is the predicted velocity, and \(v_t\) is the ground-truth velocity.

\textbf{Overall objective.}
In contrast to previous jewelry-specific objectives, which explicitly improve product consistency, we also employ the standard velocity prediction loss \(\mathcal{L}_{\text{MSE}}\) as in~\citet{wu2025qwen}. This loss provides global supervision over the entire try-on image and encourages the model to reconstruct the overall target distribution, including the person identity, skin tone, clothing, and background context. While $\mathcal{L}_{\text{attn}}$ and $\mathcal{L}_{\text{obj}}$ focus on preserving the structure and details of jewelry, \(\mathcal{L}_{\text{MSE}}\) serves as the primary constraint at the image-level to maintain the consistency of the person and the overall scene. The overall training objective is defined as:
\begin{equation}
    \mathcal{L}
    =
    \mathcal{L}_{\text{MSE}}
    +
    \mathcal{L}_{\text{obj}}
    +
    \mathcal{L}_{\text{attn}}
\end{equation}~\label{eqn:overall_loss}

\begin{figure*}[t]
    \centering
    \vskip -0.1in
    \includegraphics[width=\textwidth]{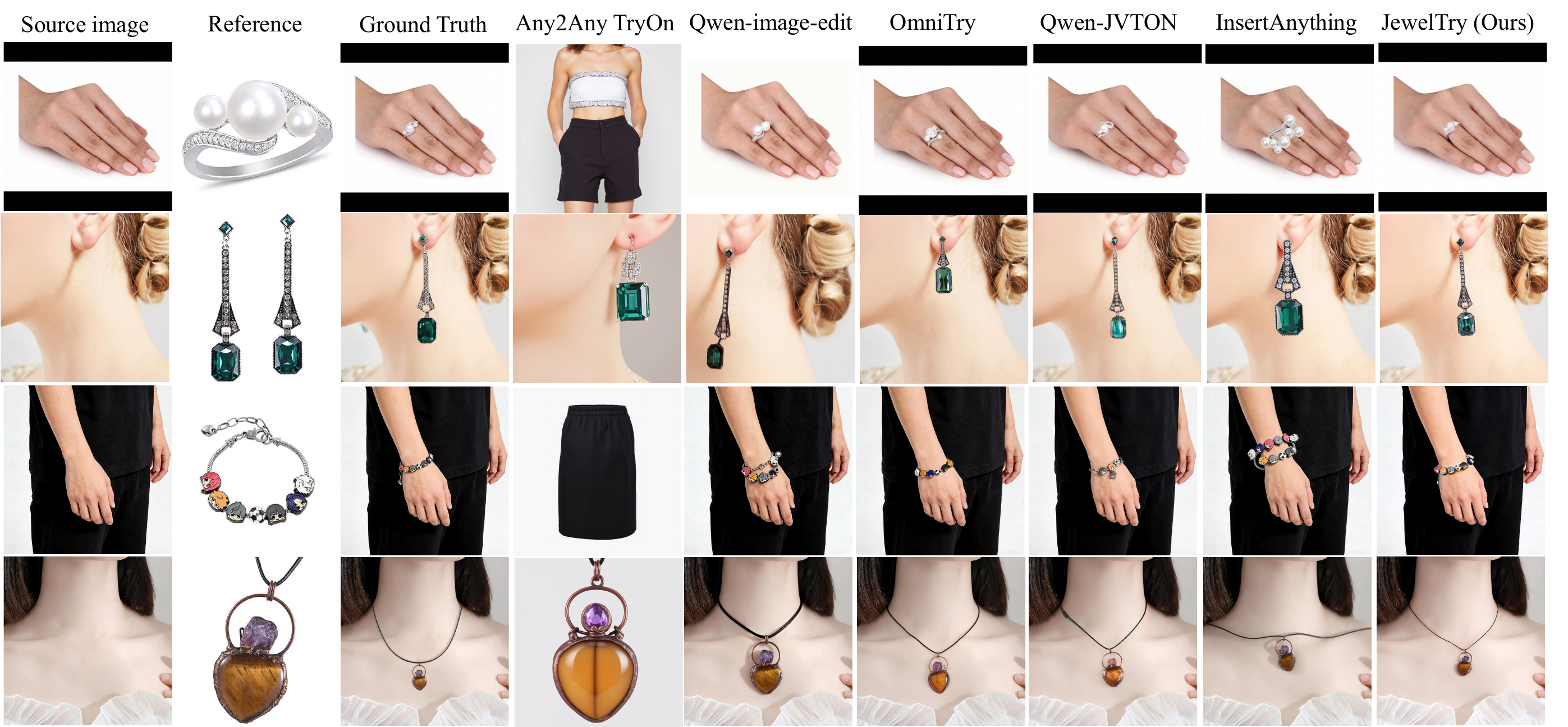}
    \caption{Qualitative comparison on JVTO-Bench test split cross four jewelry categories.}
    \vskip -0.05in
    \label{fig:comp1}
\end{figure*}

\begin{table*}[t]
  \centering
  \renewcommand{\arraystretch}{0.88}
  \setlength{\tabcolsep}{0.5mm}
  {\fontsize{8pt}{9pt}\selectfont
  \begin{tabular}{l|c|l|c|ccc|cccccc}
    \toprule
    & & &
    \multicolumn{1}{c|}{Fidelity} &
    \multicolumn{3}{c|}{Background preservation} &
    \multicolumn{6}{c}{Object consistency} \\
    \cmidrule(lr){4-4}
    \cmidrule(lr){5-7}
    \cmidrule(l){8-13}
    Method &
    Mask-free &
    Dataset &
    FID$\downarrow$ &
    DINO$_{\mathrm{p}}\uparrow$ &
    LPIPS$_{\mathrm{p}}\downarrow$ &
    SSIM$_{\mathrm{p}}\uparrow$ &
    DINO$_{\mathrm{Ref}}\uparrow$ &
    DINO$_{\mathrm{Tar}}\uparrow$ &
    CLIP$_{\mathrm{Ref}}\uparrow$ &
    CLIP$_{\mathrm{Tar}}\uparrow$ &
    IoU$\uparrow$ &
    ScaleErr$\downarrow$ \\
    \midrule

    Ground Truth
    & -- & JVTO-Bench
    & -- & -- & -- & --
    & 0.567 & -- & 0.799 & -- & -- & -- \\

    Qwen-Image-Edit
    & \checkmark & JVTO-Bench
    & 39.72 & 0.926 & 0.249 & 0.805
    & \textbf{0.617} & 0.749 & \textbf{0.820} & 0.891
    & 0.469 & 0.263 \\

    OmniTry
    & \checkmark & JVTO-Bench
    & \underline{31.23} & \underline{0.953} & \underline{0.126}
    & \textbf{0.879}
    & 0.563 & \underline{0.761} & 0.773 & \underline{0.899}
    & \textit{0.538} & \textit{0.256} \\

    Any2AnyTryOn
    & \checkmark & JVTO-Bench
    & 73.64 & 0.368 & 0.755 & 0.544
    & 0.511 & 0.457 & 0.771 & 0.785
    & 0.136 & 1.158 \\

    InsertAnything
    & $\times$ & JVTO-Bench
    & 54.47 & 0.919 & 0.156 & 0.869
    & \underline{0.579} & 0.660 & \underline{0.805} & 0.868
    & 0.409 & 0.397 \\

    Qwen-JVTON
    & \checkmark & JVTO-Bench
    & \textit{32.62} & \textit{0.950} & \textit{0.128}
    & \textit{0.867}
    & 0.525 & \textit{0.759} & 0.774 & \textit{0.898}
    & \underline{0.591} & \underline{0.208} \\

    \rowcolor{gray!10}
    \textbf{JewelTry (Ours)}
    & \checkmark & JVTO-Bench
    & \textbf{30.49} & \textbf{0.959} & \textbf{0.121}
    & \underline{0.873}
    & \textit{0.565} & \textbf{0.792} & \textit{0.801}
    & \textbf{0.914} & \textbf{0.658} & \textbf{0.169} \\
    \midrule

    Qwen-Image-Edit
    & \checkmark & OmniTry-Bench
    & -- & 0.926 & 0.223 & 0.685
    & \textbf{0.556} & -- & \textbf{0.781} & -- & -- & -- \\

    OmniTry
    & \checkmark & OmniTry-Bench
    & -- & 0.992 & \underline{0.017} & \underline{0.960}
    & \textit{0.517} & -- & \textit{0.761} & -- & -- & -- \\

    Any2AnyTryOn
    & \checkmark & OmniTry-Bench
    & -- & 0.539 & 0.636 & 0.429
    & 0.345 & -- & 0.708 & -- & -- & -- \\

    InsertAnything
    & $\times$ & OmniTry-Bench
    & -- & \textit{0.993} & \textbf{0.012} & \textbf{0.987}
    & 0.502 & -- & 0.760 & -- & -- & -- \\

    Qwen-JVTON
    & \checkmark & OmniTry-Bench
    & -- & \underline{0.996} & 0.034 & 0.928
    & 0.471 & -- & 0.717 & -- & -- & -- \\

    \rowcolor{gray!10}
    \textbf{JewelTry (Ours)}
    & \checkmark & OmniTry-Bench
    & -- & \textbf{0.997} & \textit{0.033} & \textit{0.929}
    & \underline{0.540} & -- & \underline{0.762} & -- & -- & -- \\
    \bottomrule
  \end{tabular}
  }
  \caption{Quantitative evaluations between our method and other baselines. \textbf{First} , \underline{Second}, and
 \textit{Third} indicate the best, second best, and third best performance respectively.}
  \label{tab:compare}
  \vspace{-0.15in}
\end{table*}
\section{Experiment and Results}

\subsection{Evaluation metrics}
We evaluate jewelry virtual try-on performance from three perspectives: (1) image fidelity; (2) background preservation; and (3) object consistency.
We measure the fidelity of generated images using Fréchet Inception Distance (FID), computed between the generated try-on images and the ground-truth target images. FID evaluates the distributional similarity between generated and real images and serves as a measure of image realism.
Following~\cite{feng2025omnitry}, we assess how well the non-jewelry regions are preserved after try-on. We compute DINO$_{\text{p}}$~\cite{zhang2022dino}, LPIPS$_{\text{p}}$~\cite{zhang2018perceptual}, and SSIM$_{\text{p}}$~\cite{wang2004image} on the masked-out non-jewelry regions between the generated image and the target image. 
To evaluate whether the generated jewelry faithfully matches the reference jewelry, we compute DINO$_{\text{Ref}}$ and CLIP$_{\text{Ref}}$ between the cropped generated jewelry region and the reference jewelry image. In addition, we compute DINO$_{\text{Tar}}$ and CLIP$_{\text{Tar}}$ between the cropped generated jewelry region and target jewelry region. These metrics capture both structural and semantic consistency of the generated jewelry. To assess scale and placement accuracy, we compute the IoU between the predicted jewelry region and the corresponding ground-truth jewelry region in the target image. A higher IoU indicates better alignment in both object scale and spatial placement. We also report scale error rate by calculating the log ratio between predict jewelry and ground-truth jewelry via bounding box, the lower scale error rate indicating the better scale faithfulness.

\subsection{Comparison}
As we are the first to focus on mask-free scale-aware jewelry VTON, no prior method shares our exact setting. To enable a comprehensive and fair comparison, we evaluate against five baselines spanning two groups. \textbf{Zero-shot baselines} use off-the-shelf models without any adaptations: (1) Qwen-Image-Edit~\cite{wu2025qwen}, an image-editing model supporting multi-image prompts; (2) OmniTry~\cite{feng2025omnitry}, a mask-free VTON model supporting jewelry categories; (3) Any-to-any TryOn~\cite{guo2025any2anytryon}, a mask-free garment VTON model with free-text prompts; and (4) InsertAnything~\cite{song2026insert}, a mask-guided object-insertion
model. \textbf{Trained-on-bench baselines} are fine-tuned on the JVTO-Bench training split under the same dataset as JewelTry: (5) Qwen-JVTON, a Qwen-Image-Edit model LoRA-fine-tuned on JVTO-Bench with scale information embedded into the text prompt, which serves as our most directly comparable baseline. We exclude ShiningYourself~\cite{miao2025shining} and SparklingTogether~\cite{xu2026sparkling}, due to the lack of publicly available training data and implementation details, which would preclude a fair comparison.

\textbf{Dataset.}
We evaluate JewelTry on two datasets: the JVTO-Bench test split and the jewelry subset of OmniTry-Bench~\cite{feng2025omnitry}. The JVTO-Bench contains 375 samples for evaluation, each consisting of a person image, a reference jewelry image, and a ground truth, with scale annotation and caption. It covers four jewelry categories, including rings, earrings, necklaces, and bracelets, with 59, 104, 106, and 106 product-scale annotations, respectively. In addition, we construct a jewelry subset from OmniTry-Bench, comprising 300 paired object-person samples in total without ground-truths and scale annotations. This subset includes 15 independent person images and 5 clean-background reference images for each jewelry category.

\textbf{Qualitative Comparison.}
Figure~\ref{fig:comp1} compares try-on results across jewelry categories with different structures and wearing regions. The general image-editing model Qwen-Image-Edit fails to preserve the source image consistency, demonstrating the difficulty of directly applying general editing models to jewelry VTON. In contrast, OmniTry and Qwen-JVTON better preserve the source image, but tend to lose jewelry details and structural consistency. Any2AnyTryOn is primarily designed for garment VTON and does not generalize well to off-the-shelf jewelry objects. InsertAnything, as a mask-guided object insertion method, achieves relatively strong object preservation, but produces weaker integration with the human body. In contrast, our method has better performance on preserving jewelry scale and structural details, which produces more realistic and visually coherent jewelry virtual try-on results.

\textbf{Quantitative Comparison.}
Table~\ref{tab:compare} reports the results of JVTO-Bench and OmniTry-Bench. In JVTO-Bench, JewelTry achieves the best score among all baselines compared on FID, DINO$_{\text{p}}$, LPIPS$_{\text{p}}$, DINO$_{\text{Tar}}$,
CLIP$_{\text{Tar}}$, IoU and ScaleErr, which indicate JewelTry achieves better fidelity, background preservation, jewelry object consistency, and scale accuracy. Although JewelTry does not achieve the highest DINO$_{\text{Ref}}$ and CLIP$_{\text{Ref}}$ scores, its scores remain close to the ground truth. Notably, these scores approaching 1 are not necessarily desirable, as they may indicate near-direct copying of the reference image rather than realistic re-rendering (See the bracelet and ring examples in Figure~\ref{fig:comp1}). Target-supervised metrics and jewelry scale are inapplicable on OmniTry-Bench dataset. We drop the scale tokens for JewelTry during inference and obtains the best DINO$_{\text{p}}$ and third-best LPIPS$_{\text{p}}$/SSIM$_{\text{p}}$. Qwen-Image-Edit leads the reference-based object-consistency score there, but at the cost of person preservation. Overall, JewelTry achieves a balance among visual fidelity, background preservation, object consistency and scale accuracy. We provide more experimental results and implementation details in our supplementary material.

\begin{table}
  \centering
    \renewcommand{\arraystretch}{1}
    \renewcommand{\tabcolsep}{0.7mm}
{\fontsize{8pt}{9pt}\selectfont
  \begin{tabular}{l|ll|lll}
    \toprule
    Method &DINO$_{\text{p}}$$\uparrow$ & LPIPS$_{\text{p}}$$\downarrow$ &  DINO$_{\text{ref}}\uparrow$  & DINO$_{\text{tar}}\uparrow$ & IoU$\uparrow$ \\
    \midrule
    Qwen-JVTON & 0.950 & 0.128 & 0.525 & 0.759 & 0.591\\
    +SDAttn & 0.953 & 0.121 & 0.559 & 0.760 & 0.589 \\
    +SDAttn \& SA & 0.958 & 0.124 & 0.545 & 0.772 & 0.634\\
    +SDAttn\&SA\&$\mathcal{L}_{\text{attn}}$ & 0.957  & 0.122  & 0.559 & 0.788  & 0.635\\ 
    \midrule
    JewelTry & \textbf{0.959} & \textbf{0.121}  &  \textbf{0.565} &  \textbf{0.792} &  \textbf{0.658} \\
    \bottomrule
  \end{tabular}}
  \caption{Ablation study for JewelTry on JVTO-Bench, where SA stands for scale adapter and SDAttn stands for single direction condition attention.}
  \label{tab:ablation}
\end{table}

\begin{figure}[t]
    \centering
    \includegraphics[width=\linewidth]{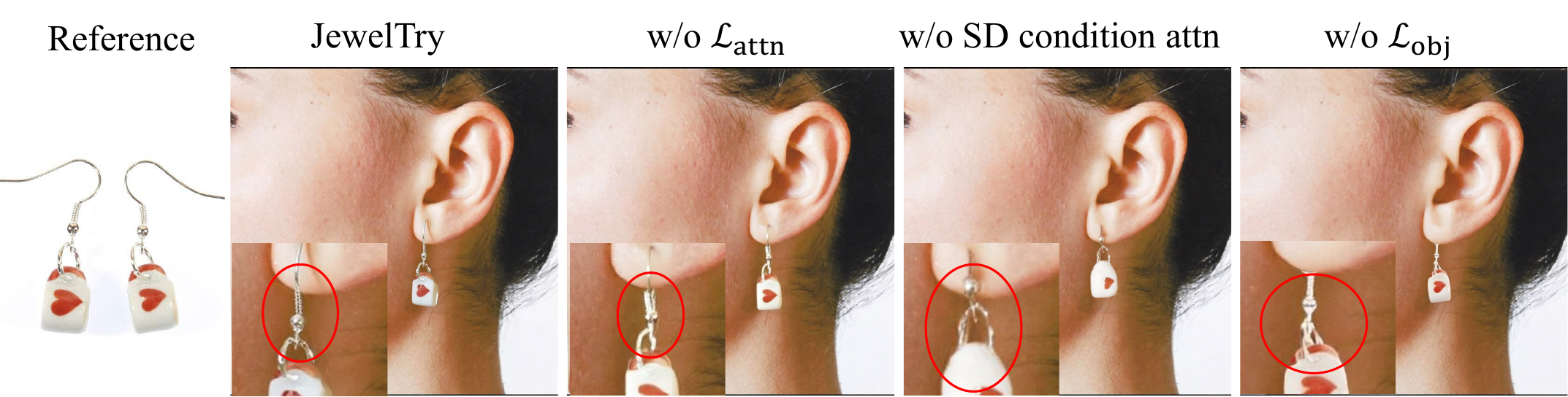}
     \caption{The visual comparisons of our models with different module configurations for jewelry consistency.}
    \label{fig:ablation}
    \vskip -0.15in
\end{figure}

\subsection{Ablation study}
We conduct ablation studies by progressively adding each proposed component to the baseline, Qwen-JVTON (Table~\ref{tab:ablation}), and assess whether each component improves the specific aspect it targets. Introducing single-directional condition attention (SDAttn) primarily targets reference consistency and yields the largest single gain in DINO$_{\text{ref}}$, indicating better preservation of reference jewelry structure. Since Qwen-JVTON is trained with scale information embedded in the VLM's inputs, it exhibits a degree of scale awareness. Adding the scale adapter further enhances scale-awareness in the model, and accordingly increases IoU; We observe a small decrease in DINO$_{\text{ref}}$ and small increases in DINO$_{\text{tar}}$, which is expected as inference variance.
The $\mathcal{L}_{\text{attn}}$ further improves target-region consistency by sharpening alignment between jewelry condition tokens and the target try-on region. Finally, adding $\mathcal{L}_{\text{obj}}$ yields the full JewelTry model, which attains the best on object-consistency metric while maintaining background preservation. In addition, $\mathcal{L}_{\text{obj}}$ enhances the correctness of jewelry placement with an increased IoU. Figure~\ref{fig:ablation} provides a qualitative comparison which provides a better interpretation.
Without the attention refinement loss, the model fails to fully preserve fine-grained jewelry details, leading to slight structural changes in the earring. Removing the single-directional condition attention further degrades coarse-level object consistency, producing jewelry with distorted overall structure. Similarly, removing the object consistency loss weakens reference preservation and results in noticeable deviations from the earring structure. In contrast, the full JewelTry model better maintains both the coarse geometry and fine-grained details of the reference jewelry, demonstrating the effectiveness of these components for object consistency.

\begin{figure}[t]
    \vskip -0.1in
    \centering
    \includegraphics[width=\linewidth]{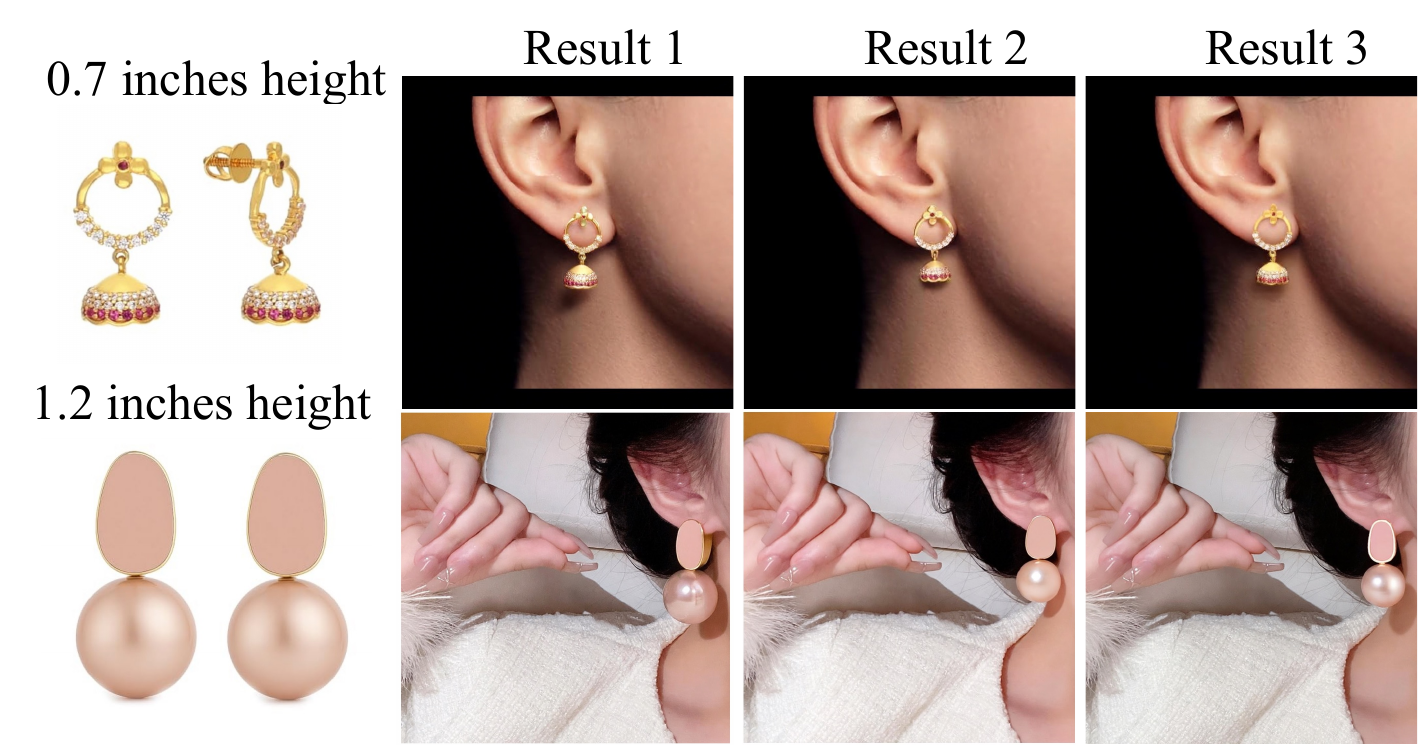}
    \vskip -0.05in
     \caption{Despite implicit size control in JewelTry, repeated generations converge to similar relative jewelry scales.}
    \label{fig:scale_control}
\end{figure}

\begin{figure}[t]
    \vskip -0.1in
    \centering
    \includegraphics[width=\linewidth]{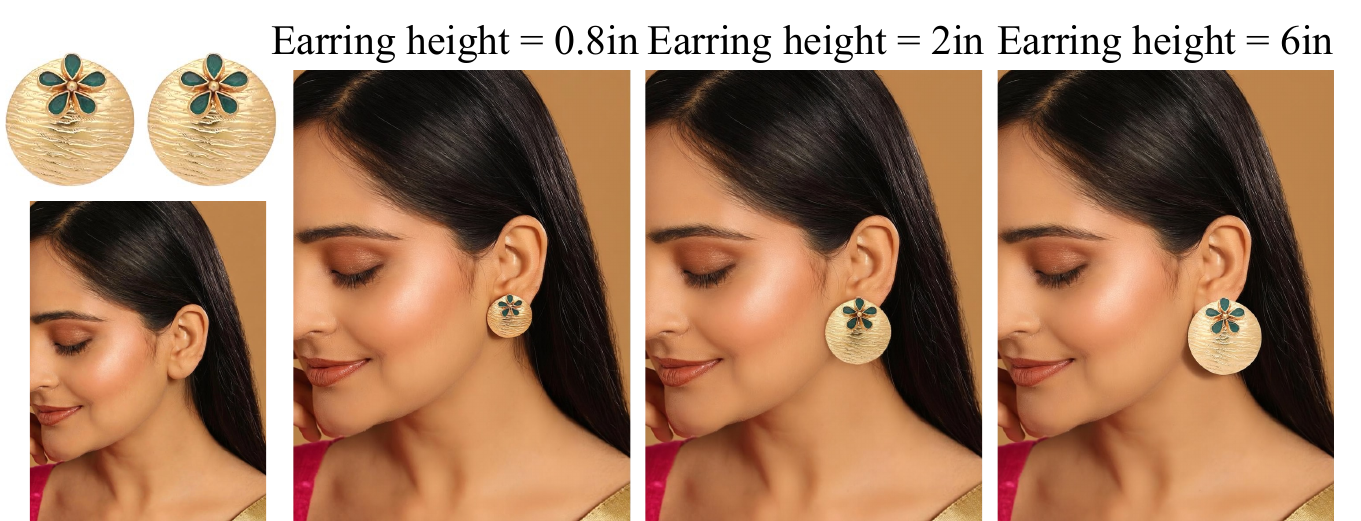}
    \vskip -0.05in
     \caption{JewelTry enables relative size control but may fail at extreme scales due to its reliance on implicit in-context learning and the scale bias inherent in real-world jewelries.}
    \label{fig:scale_control2}
    \vskip -0.2in
\end{figure}

\subsection{Discussion and limitation}
As the first mask-free scale-aware jewelry VTON method, JewelTry uses a scale token as implicit condition to model the relationship between jewelry and human anatomy. While this improves scale consistency, it does not guarantee precise absolute scale control. Repeated inference with the same jewelry image, person image, and scale input may produce slight variations in results (Figure~\ref{fig:scale_control}). Moreover, JewelTry mainly provides \emph{relative} rather than \emph{absolute} scale control: increasing the scale token consistently enlarges the generated jewelry, but the visual size may not exactly match the extreme large or small dimension, e.g., a 6-inch earring may not be rendered at a perceptually exact 6-inch scale (Figure~\ref{fig:scale_control2}). This limitation may arise from data bias, as extreme scales are rare in real-world jewelry distributions, and from the inherent difficulty of mapping physical dimensions to image-space geometry without explicit geometric supervision. 

\section{Conclusion}
In this paper, we address mask-free scale-aware jewelry virtual try-on, a challenging setting where visual realism depends critically on object scale and fine-grained structural preservation. We introduce JVTO-Bench, a dataset with triplet samples and product-scale annotations across four major jewelry categories. Building on this, we propose JewelTry, a mask-free framework that incorporates product scale through a scale adapter and improves jewelry fidelity via single-directional condition attention and an attention refinement loss. These designs preserve jewelry appearance and structure while rendering it at scale-aware without relying on masks.  Extensive experiments show that JewelTry achieves a balance among visual fidelity, background preservation, object consistency, and scale accuracy, establishing a strong baseline for mask-free, scale-aware jewelry virtual try-on for future research.

\bibliography{aaai2027}


\end{document}


\maketitle

\tableofcontents

\section{JVTO-Bench}

JVTO-Bench is a real-world jewelry virtual try-on dataset covering four major jewelry categories: rings, earrings, necklaces, and bracelets. The dataset contains approximately 23k samples with diverse jewelry items and pose.  JVTO-Bench provides triplet samples including ground-truth try-on image, try-off person image, and reference jewelry image. JVTO-Bench also provides real-world product 2D scale in inches, product type, how jewelry object interact with the person image caption, and target object bounding box in try-on image annotations.

Each data sample includes a reference product image captured on a white background, a try-on image showing the jewelry worn by a model, a corresponding try-off image without the jewelry, an object mask indicating the jewelry region in the try-on image, and product type, $H\times W$ scale (in inches), and caption annotations. JVTO-Bench is designed to support the development and evaluation of jewelry virtual try-on methods across diverse product types, visual appearances, and market sources.

Table~\ref{tab:JVTO_bench_stat} summarizes the statistics of JVTO-Bench. To prevent product-level overlap between the training and test sets, we randomly reserve 150 products from each of the ring, earrings, necklace, and bracelet categories before constructing the training split. After data filtering, the final test set contains 59 ring products, 104 earring products, 106 necklace products, and 106 bracelet products. The ring category is treated separately because its scale annotations are more heterogeneous: many ring listings either omit explicit dimension information or provide multiple size variants. We therefore retain the 59 valid ring samples as a separate, potentially biased test subset, while keeping the other three categories approximately balanced at around 100 samples each. Ground-truth try-on images in JVTO-Bench have resolutions ranging from 500 to 2,000 pixels, and images with a minimum spatial resolution below 500 pixels are discarded. Representative samples are provided in the dataset supplementary material.

\begin{table}[ht]
    \centering
    \scalebox{0.8}{
    \begin{tabular}{c|cccc|c}
        \toprule
         & Ring & Earrings & Necklace & Bracelet & Total  \\
         \midrule
         Train split & 2908	&12149 & 4586 &4019	&23662\\
         Test split & 59	&104&	106	&106	&375 \\
         \bottomrule
    \end{tabular}
    }
    \caption{JVTO-Bench Statistics.}
    \label{tab:JVTO_bench_stat}
\end{table}

\begin{figure}[ht]
    \centering
    \includegraphics[width=\linewidth]{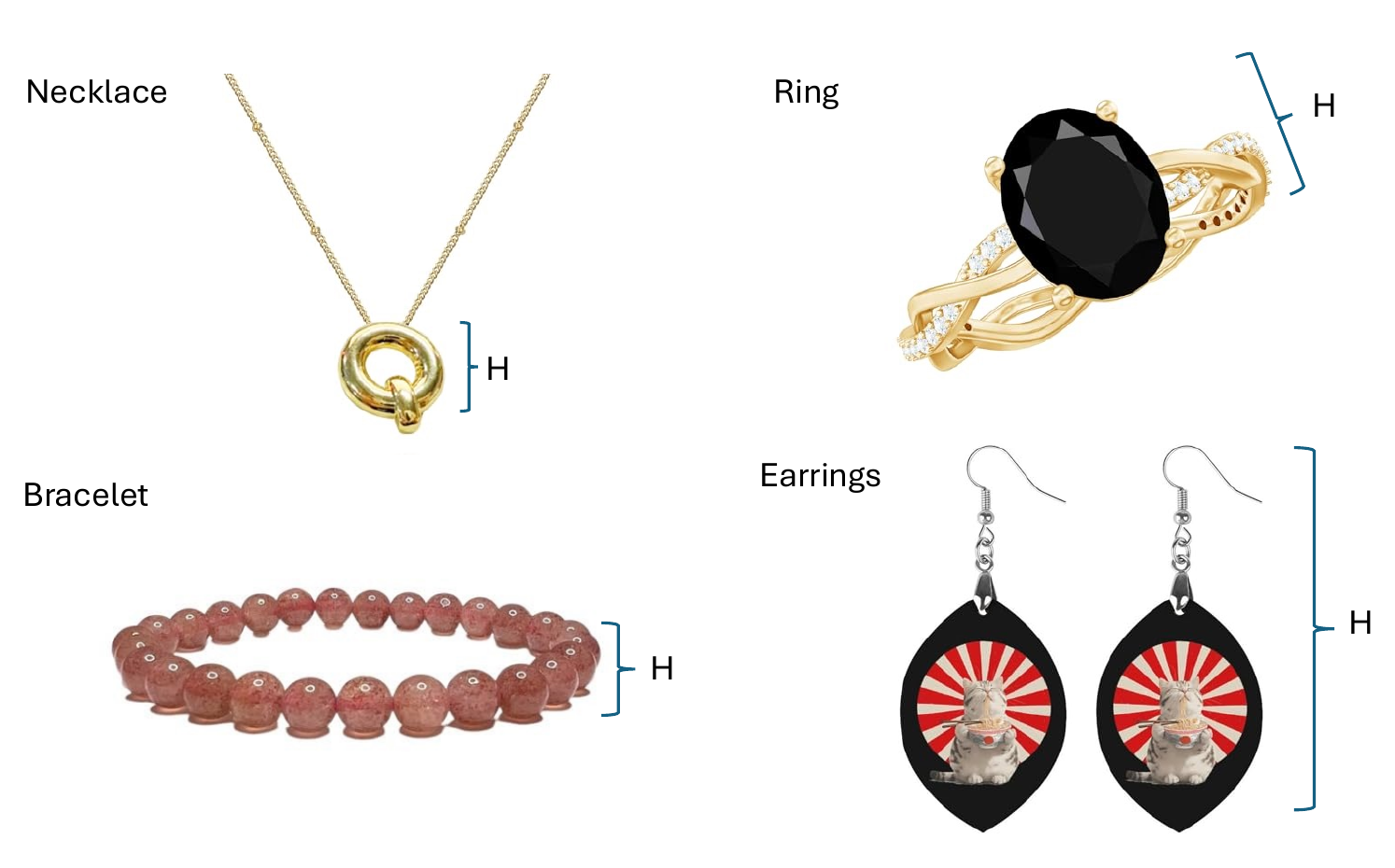}
    \caption{Product-scale annotations across different jewelry categories. Although jewelry items are 3D objects, we annotate category-specific 2D physical dimensions that are commonly provided in online shopping, such as length, width, diameter, or pendant size. This design follows real-world e-commerce practice, where sellers typically describe the most visually relevant dimensions for each jewelry category. }
    \label{fig:JVTO_bench_scale}
\end{figure}

\begin{figure}[ht]
    \centering
    \includegraphics[width =\linewidth]{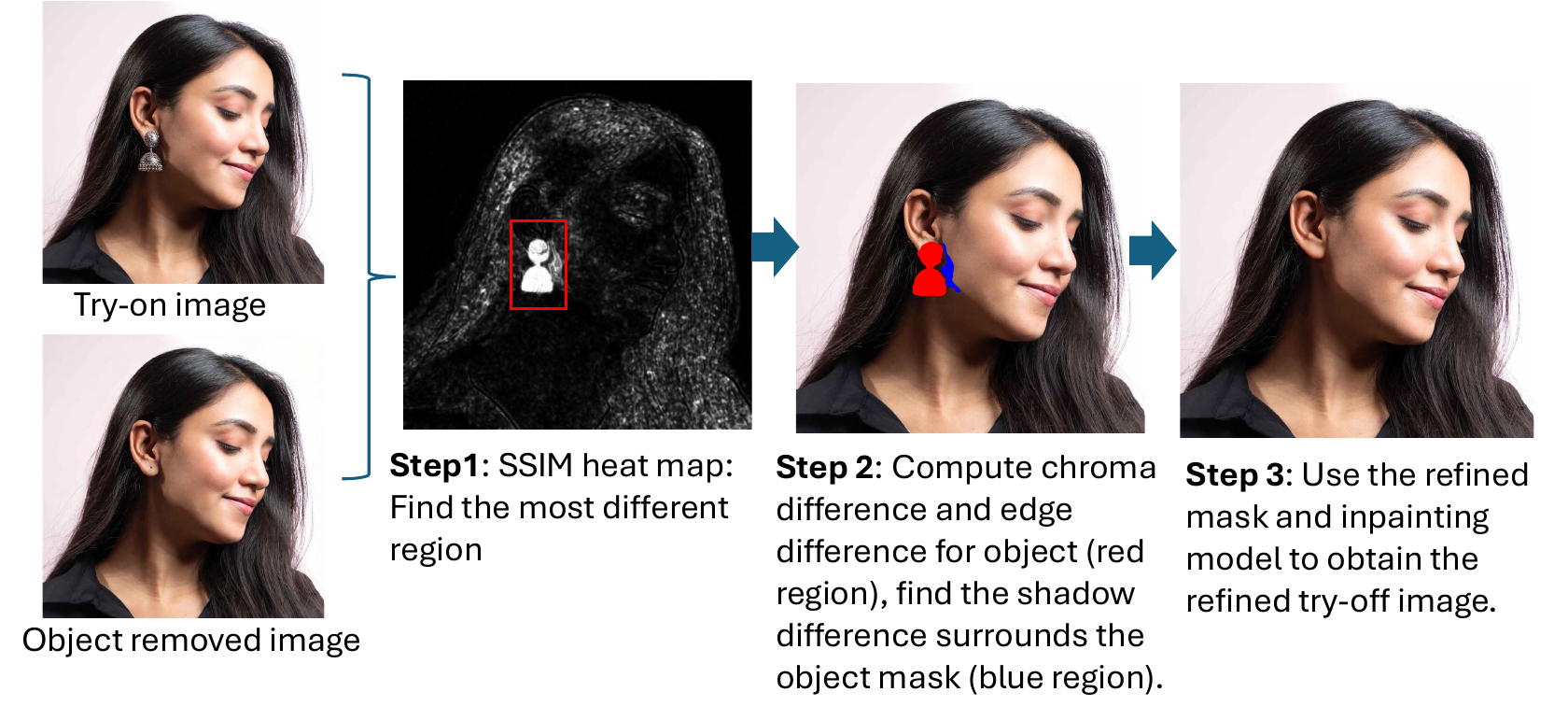}
    \caption{Our refined try-off image processing pipeline. We further refine try-off image for those low quality Qwen-Object-Remover outputs.}
    \label{fig:refined}
\end{figure}

\begin{figure}[ht]
    \centering
    \includegraphics[width=\linewidth]{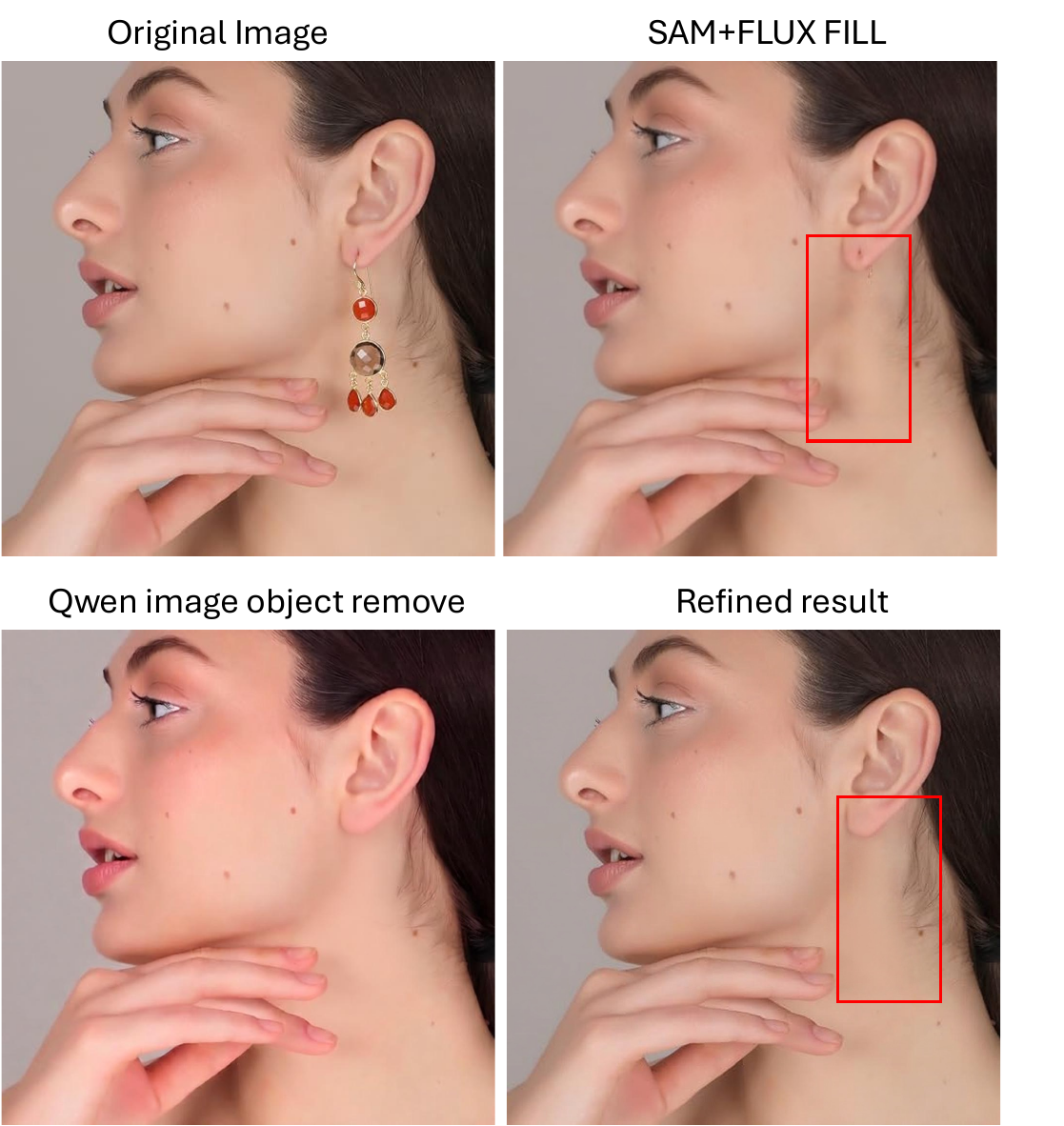}
    \caption{Comparison of different object removal methods. Our refined try-off images exhibit more natural visual quality and better suppress residual shadows and artifacts caused by the removed jewelry.}
    \label{fig:JVTO_remove}
\end{figure}

Since jewelry is a rigid, solid object with fixed geometry, its proportions therefore remain constant, so a single dominant dimension is sufficient to characterize its overall scale. As illustrated in Figure~\ref{fig:JVTO_bench_scale}, we define category-specific scale annotations according to the structural characteristics of each jewelry type. For earrings, $H$ denotes the overall vertical height of the earring; for necklaces, H corresponds to the pendant height. For bracelets and rings, whose worn appearance is mainly determined by band thickness, we define H as the maximum width of the bracelet or ring band. This unified scale representation enables consistent modeling of jewelry size across categories. Since jewelry items are rigid and their perceived size is largely determined by a single dominant dimension, we adopt a unified one-dimensional scale representation and use only the $H$ dimension during training. In JVTO-Bench, however, we provide 2D scale for future research, where $W$ dimension is perpendicular to $H$ dimension indicated in Figure~\ref{fig:JVTO_bench_scale}.

Obtaining high-quality try-off images is particularly challenging for jewelry VTON. Since jewelry items often cast shadows or introduce subtle local appearance changes, simply combining object detection with an inpainting model~\cite{feng2025omnitry} may fail to fully remove the jewelry-related regions. Object removal models such as Qwen-Object-Remover can more effectively eliminate both the jewelry and its shadow, producing more consistent try-off images; however, they may also introduce color shifts or unintended changes in some cases. To improve try-off quality, we further refine the object removal results by identifying changed regions between the original image and the object-removed image using SSIM. We then use the resulting difference region as a refined mask and apply an additional inpainting step to better preserve the surrounding appearance (see Figure~\ref{fig:refined}). Figure~\ref{fig:JVTO_remove} shows a qualitative comparison of try-off image generation using different removal strategies. However, we observe that the SSIM-based mask is not robust across all images. Therefore, we manually apply this refinement step only to Qwen object-removal results that exhibit obvious quality degradation.

\begin{figure*}[t]
    \centering
    \begin{subfigure}[t]{0.48\linewidth}
        \centering \includegraphics[width=\linewidth]{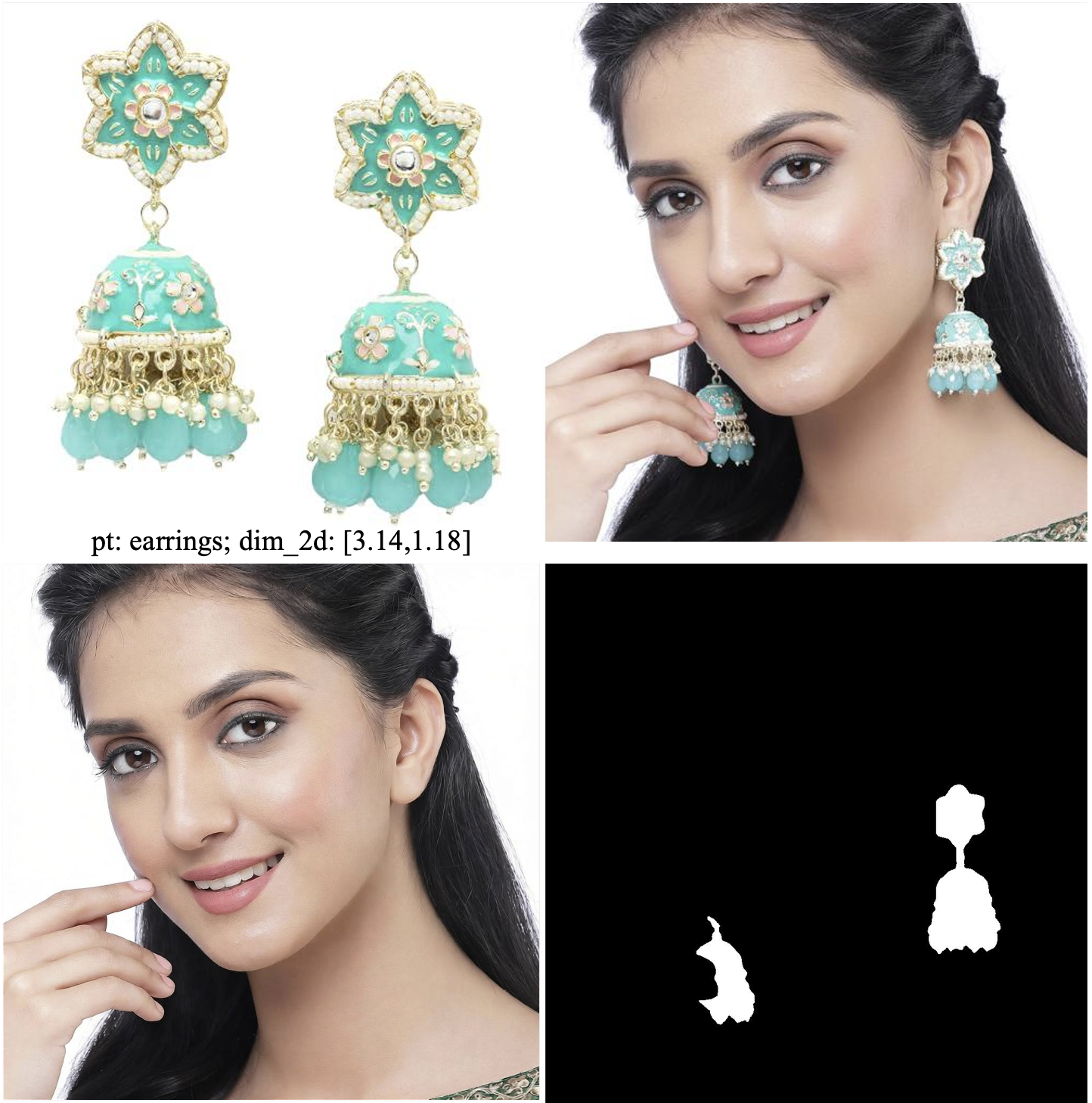}
        \caption{Earrings sample.}
    \label{fig:JVTO_bench_earrings}
    \end{subfigure}
    \hfill
    \begin{subfigure}[t]{0.48\linewidth}
        \centering
        \includegraphics[width=\linewidth]{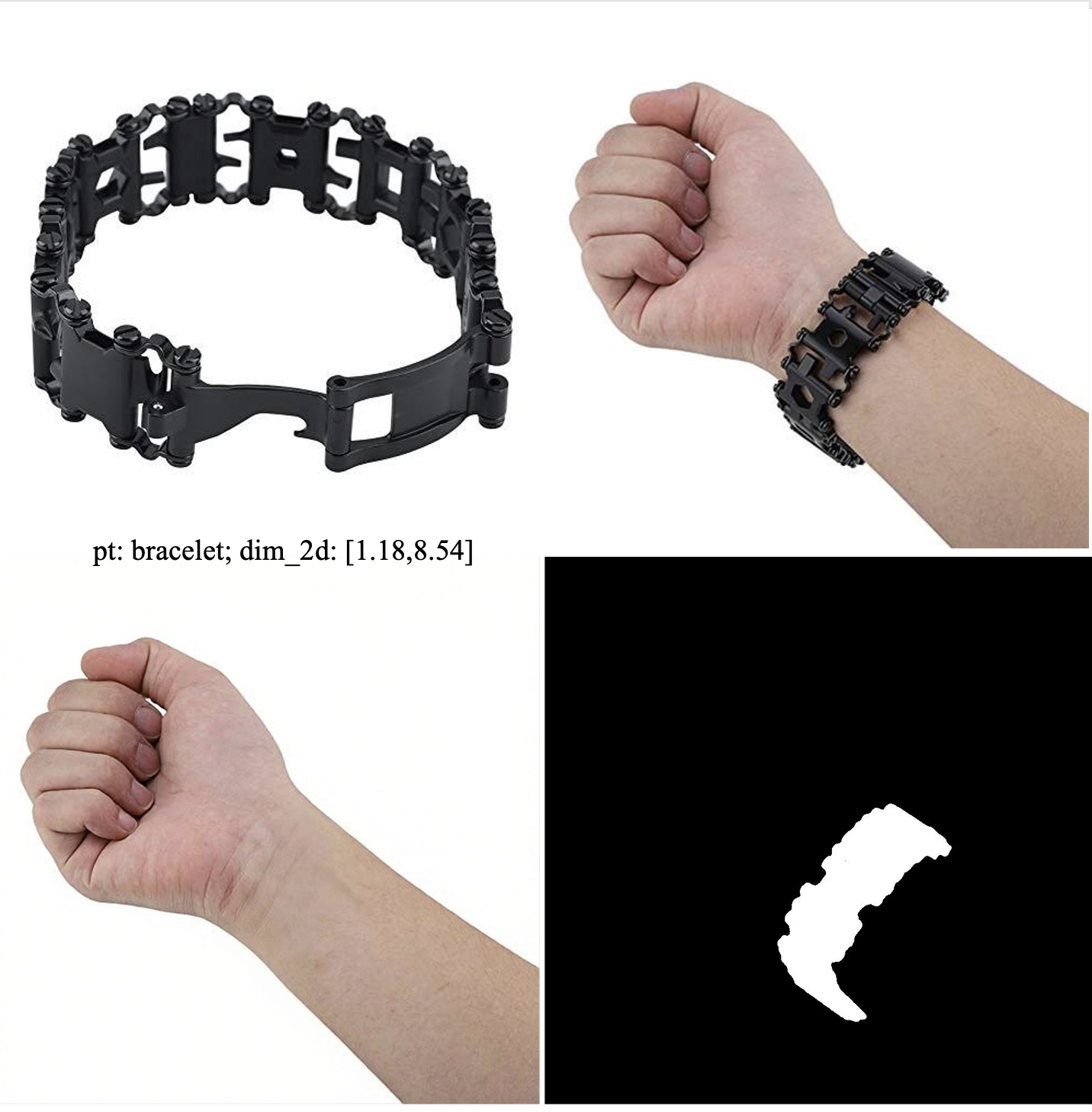}
        \caption{Bracelet sample.} \label{fig:JVTO_bench_bracelet}
    \end{subfigure} \vspace{0.5em}
    \begin{subfigure}[t]{0.48\linewidth}
        \centering
        \includegraphics[width=\linewidth]{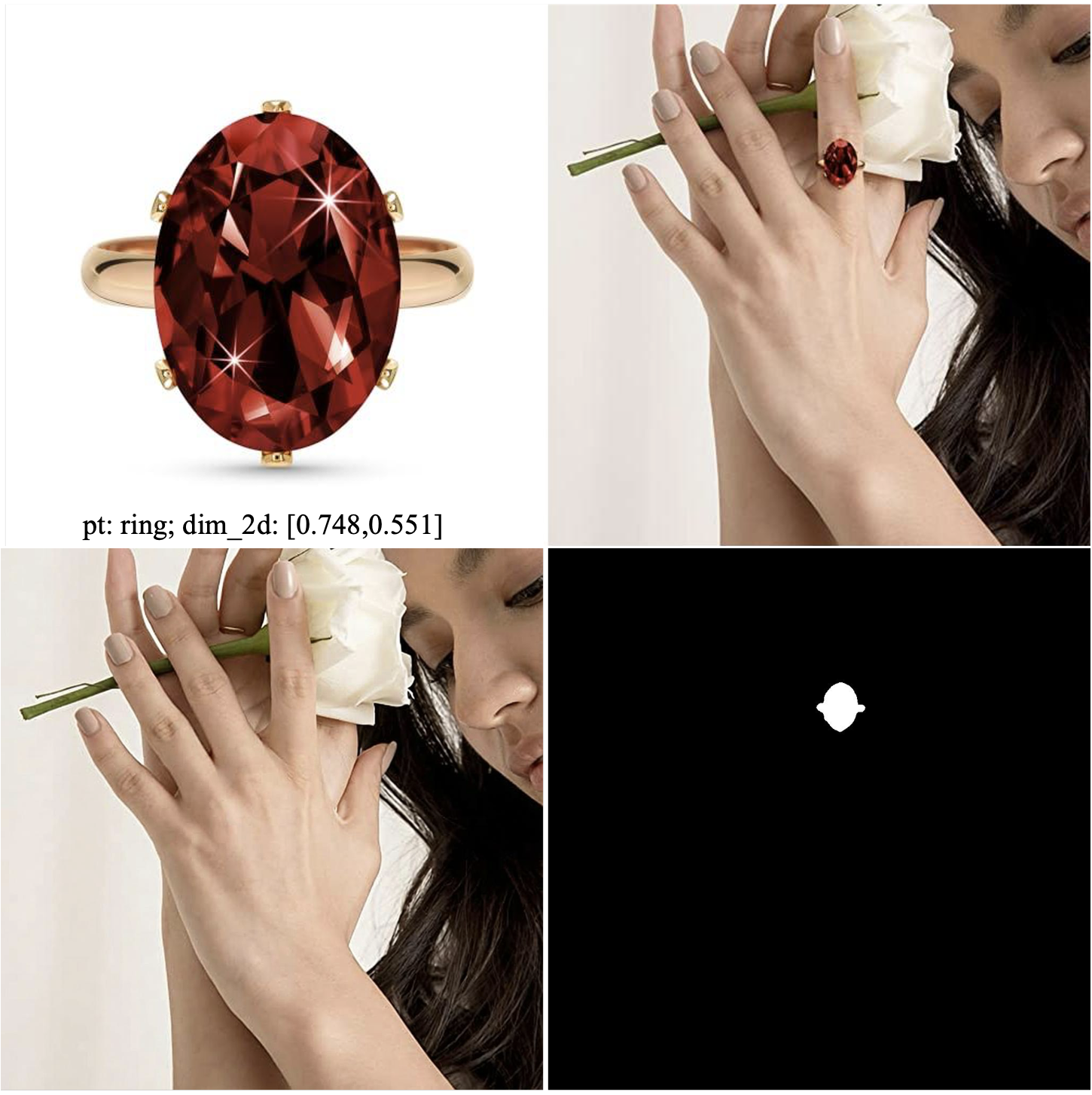}
        \caption{Ring sample.} \label{fig:JVTO_bench_ring}
    \end{subfigure}
    \hfill
    \begin{subfigure}[t]{0.48\linewidth}
        \centering
        \includegraphics[width=\linewidth]{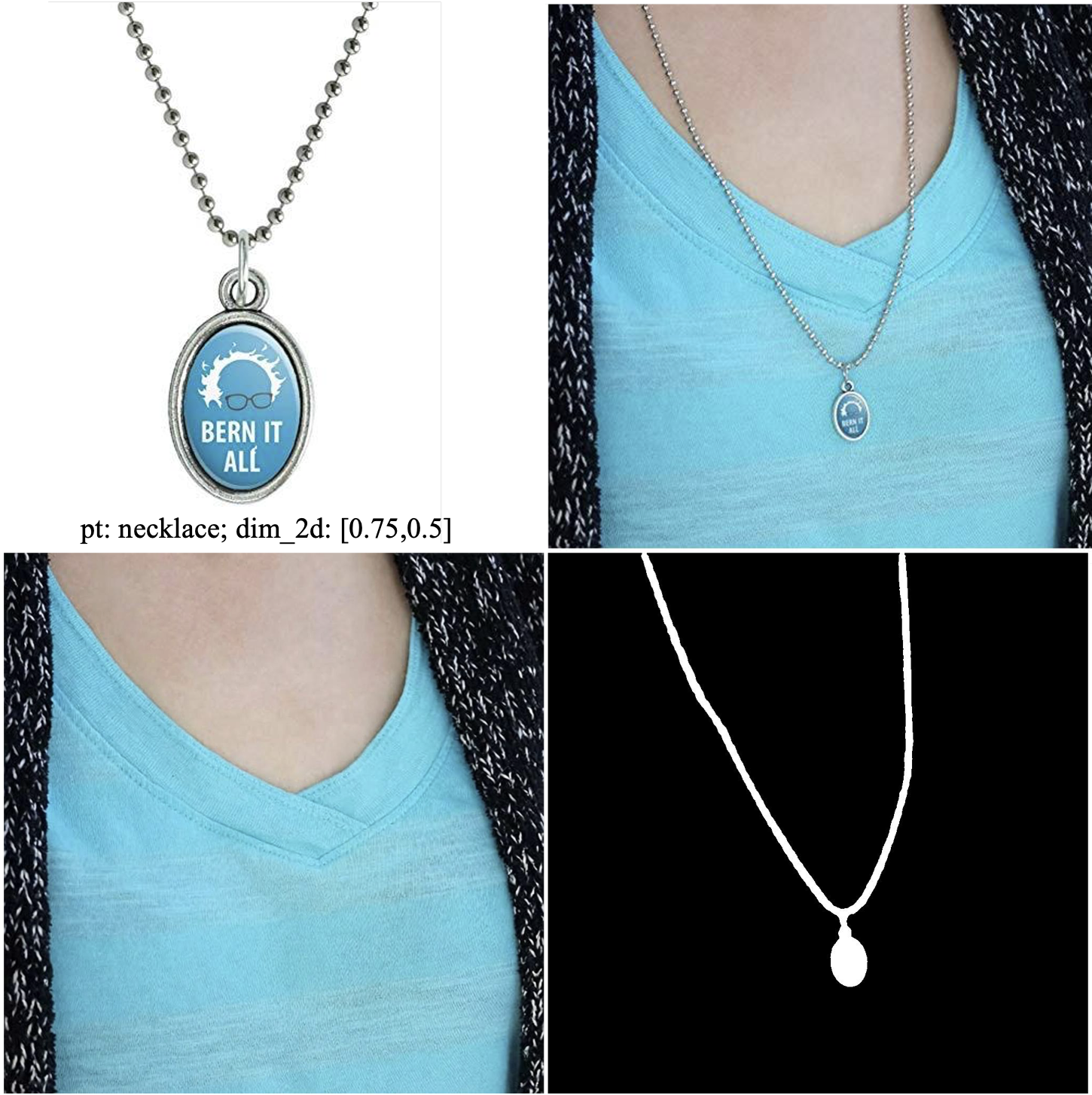}
        \caption{Necklace sample.} \label{fig:JVTO_bench_necklace}
    \end{subfigure}
    \caption{Examples from JVTO-Bench across four jewelry categories: earrings, bracelet, ring, and necklace.} \label{fig:JVTO_bench_examples}
\end{figure*}

We provide an example sample of JVTO-Bench in Figure~\ref{fig:JVTO_bench_examples}.
\begin{table}[ht]
    \centering
    \scalebox{0.8}{
    \begin{tabular}{c|cccc|c}
        \toprule
         & Ring & Earrings & Necklace & Bracelet & Total  \\
         \midrule
         Train split & 2908	&12149 & 4586 &4019	&23662\\
         Test split & 59	&104&	106	&106	&375 \\
         \bottomrule
    \end{tabular}
    }
    \caption{JVTO-Bench Statistics.}
    \label{tab:JVTO_bench_stat}
\end{table}

\section{Implementation details}

\subsection{Training}
JewelTry is built upon Qwen-Image-Edit-2511~\cite{wu2025qwen}. We fine-tune the model using LoRA adapters with rank 16 and LoRA alpha 16 inserted into the self-attention blocks ($q,k,v,out$). Training is performed on the JVTO-Bench training split using 8 NVIDIA H200 GPUs for 200k iterations with a batch size of 8 and a learning rate of $5\times10^{-5}$.

During training, the person image and target try-on image are resized to $1024\times1024$, while the reference jewelry image is resized to $768\times 768$. The proposed scale adapter first projects the numerical scale value into a 256-dimensional embedding through a linear layer followed by a SiLU activation. The resulting embedding is then combined with a learnable jewelry-category embedding and further projected to the 3584-dimensional text-token space using another SiLU-activated projection layer. To preserve the model's ability to perform inference without scale annotations, we randomly drop the scale token with a probability of 25\% during training. 

For the attention refinement loss, object attention maps are extracted from the $51^{\mathrm{st}}$ to $60^{\mathrm{th}}$ MMDiT transformer blocks, where the attention responses exhibit the strongest correlation with the target jewelry region (see examples in Figure~\ref{fig:attn_refine_vis} and Figure~\ref{fig:attn_refine_vis2}). 
To obtain an attention-based object soft mask, we extract the cross-attention maps, average them across multiple attention heads and feature dimensions, and then normalize the resulting map to the range $[0,1]$. We also involve a timestep binary gate in attention refinement loss, see Section~\ref{sec:b_gate} for more details.

Since we follow qwen-image-edit and use Qwen-VL-2.5 as the VLM to encode text prompt in JewelTry, which has reasoning ability. We use caption extracted by Claude (annotations provided by JVTO-Bench) as text prompt with additional information of product scale, e.g., ``The person from image 1 is wearing earrings from image 2. Earring height: 3.8 inches. '' for both training JewelTry and Qwen-JVTON model. To fine-tune the Qwen-image-edit model for Qwen-JVTON on JVTO-Bench dataset, we inject LoRA parameter into the self-attention block with rank 16, and we follow the exact same setting as used in JewelTry on 8 NVIDIA H200 GPUs for 200k iterations with a batch size of 8 and a learning rate of $5\times10^{-5}$ for 200k iterations.

\subsection{Inference}
For Qwen-Image-Edit-2511, Qwen-JVTON, and JewelTry, we set the classifier-free guidance (CFG) scale to 3.0 and use 50 denoising steps. For OmniTry, Any2AnyTryOn, and InsertAnything, we follow their original inference configurations.

For InsertAnything, we generate reference object masks and target inpainting masks using Grounded DINO and SAM. Since InsertAnything requires an explicit insertion mask and the mask provides direct guidance on object scale and placement. Therefore, this make mask-guided methods not directly comparable to mask-free methods. To reduce such mask-induced scale guidance while retaining a valid insertion region on JVTO-Bench, we convert the corresponding person-region mask into a rectangular bounding box and randomly enlarge the target bounding box by 20\%. For OmniTry-Bench, we directly use the target masks provided by the dataset for InsertAnything inference.

For Qwen-Image-Edit-2511 inference on JVTO-Bench, we use the same text prompts as JewelTry and Qwen-JVTON which with the scale information to ensure a fair comparison.

For inference results on OmniTry-Bench, we found that the given captions is not strongly match to the target results, which further caused Qwen-JVTON, Qwen-Image-Edit, and JewelTry to fail to preserve the person image. Therefore, we instead adopt a simplified prompt in the format: ''The person is wearing the given \{jewelry class\}``. Since OmniTry-Bench dataset has no product scale information, we drop the scale token from the scale adapter in JewelTry and also exclude scale information in the text prompt for JewelTry, Qwen-JVTON and Qwen-image-edit model.

\subsection{Seeds and randomness statement}
To ensure a fair comparison under realistic stochastic inference conditions, we generate all inference results using random seeds rather than fixing a specific seed. 

\subsection{Evaluation metrics}
Since jewelry is a rigid object with fixed geometry, its proportions remain constant, so a single dominant dimension is sufficient to characterize its overall scale. Motivated by this, we design the ScaleErr metric to evaluate scale faithfulness. Specifically, we detect the jewelry bounding box in both the generated and ground-truth try-on images, and compute the absolute log-ratio of their dominant dimension $H$, defined as

\begin{equation}
    \text{ScaleErr} =  | log( H_{\text{pred}} / H_{\text{gt}} ) |
\end{equation}

where $H_{\text{pred}}$ and $H_{\text{gt}}$ denote the dominant dimension defined in Figure~\ref{fig:JVTO_bench_scale} of the predicted and ground-truth jewelry, respectively. A value of $0$ indicates a perfect size match, and larger values indicate greater scale mismatch. The log-ratio is symmetric, penalizing equally an object rendered too large or too small by the same factor.

We use Gounded DINO~\cite{liu2023grounding} and SAM~\cite{kirillov2023segany} to detect the jewelry region, which further use the output to mask the jewelry region or calculate metrics.

\section{Quantitative comparison across jewelry categories}

\begin{figure}[t]
    \centering
    \includegraphics[width=\linewidth]{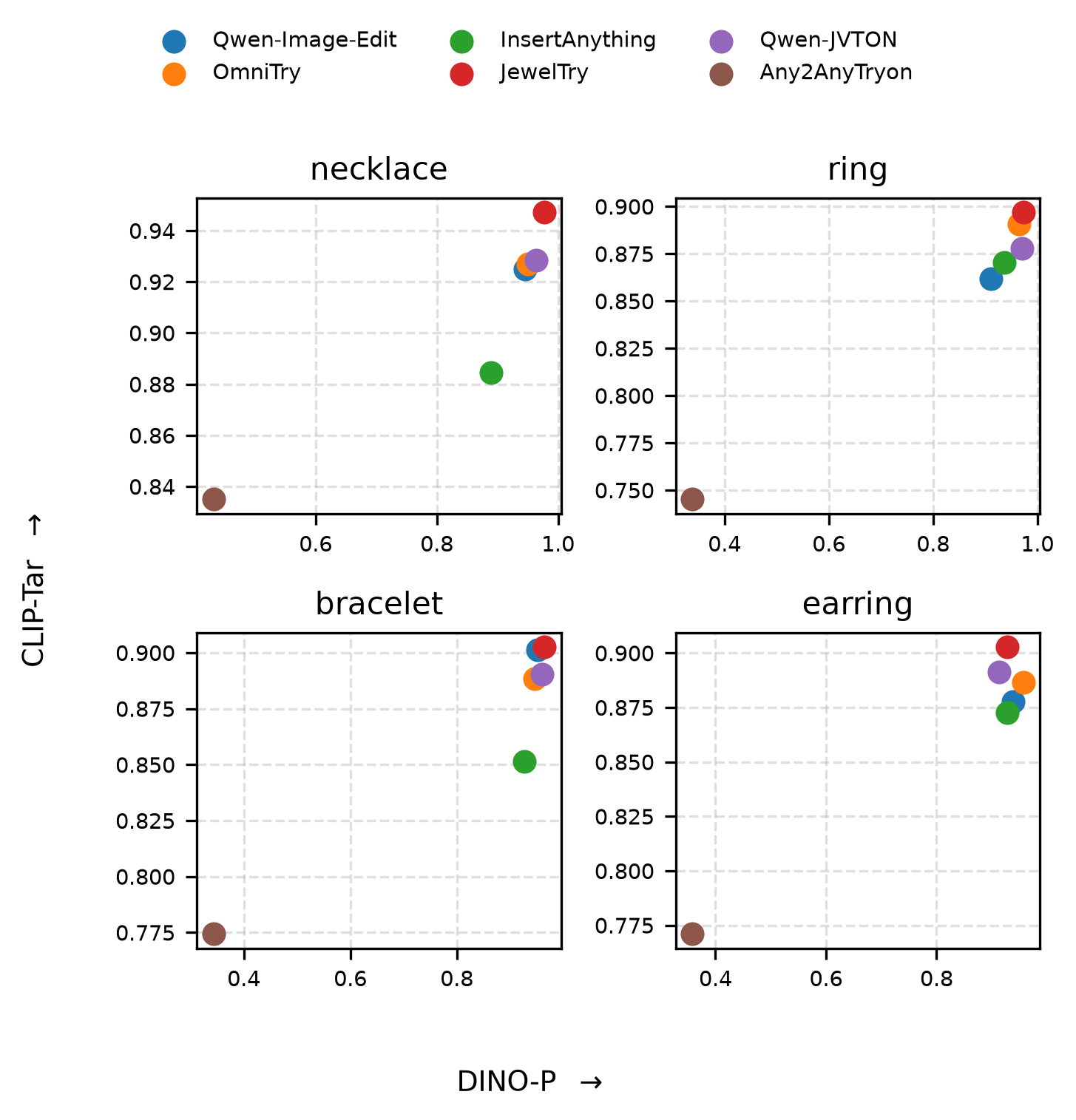}
    \caption{Quantitative comparison across four jewelry categories in JVTO-Bench. JewelTry present superior performance on maintaining object consistency and balanced background preservation cross the four jewelry categories.}
    \label{fig:compare_cate}
\end{figure}

Figure~\ref{fig:compare_cate} provides a category-wise quantitative comparison on JVTO-Bench. We report DINO$_{\text{p}}$ and CLIP$_{\text{Tar}}$, which evaluate person/background preservation and object consistency with respect to the ground-truth try-on images, respectively. Across the four major jewelry categories, JewelTry achieves the best performance on rings, necklaces, and bracelets, and ranks second on earrings, with only a small gap to the best method: $-0.0309$ on DINO$_{\text{p}}$.

These results demonstrate that JewelTry generalizes well across jewelry types with different spatial scales, body attachment regions, and structural characteristics. The strong performance on rings and bracelets indicates that our method can handle small and localized try-on regions, while the gains on necklaces suggest its effectiveness for larger accessories that require accurate placement around complex neck and upper-body regions. Although earrings remain challenging due to their tiny size, complex structure and high sensitivity to scale, and frequent occlusion by hair or face contours, JewelTry remains highly competitive. Overall, the category-wise results further validate the robustness of our scale-aware conditioning and conditional fidelity design across diverse jewelry VTON scenarios.

\section{Result variance and significance test}

We conduct significance testing (t-test) compared with each baseline on JVTO-Bench test split. Compared to the zero-shot baselines, JewelTry obtains a more balanced performance and significantly outperforms OmniTry, Qwen-image-edit, InsertAnything and Any2AnyTryOn regarding object consistency, background preservation, and scale faithfulness on metrics such as DINO$_{\text{Tar}}$ (with p-values $2.81\times 10^{-3}$, $1.1\times 10^{-2}$, $3.10 \times 10^{-22}$, and $3.87 \times 10^{-47}$, respectively), IoU (with p-values $1.19\times 10^{-13}$, $1.46\times 10^{-15}$, $1.49\times 10^{-23}$, and $1.52\times 10^{-49}$, respectively), and LPIPS$_{\text{p}}$ (with p-values $8.16\times 10^{-32}$, $7.76\times 10^{-21}$, $9.61\times 10^{-31}$, and $3.87\times 10^{-47}$, respectively). Compared to the trained-on-bench baselines, Qwen-JVTON, JewelTry also significantly outperform on object consistency, background preservation, and scale-faithfulness with p-values $8.64\times 10^{-3}$, $4.78\times 10^{-7}$, and $2.19\times 10^{-2}$ on DINO$_{\text{Ref}}$, LPIPS$_{\text{p}}$, and IoU, respectively. In terms of jewelry scale faithfulness, JewelTry significantly outperforms than all the baselines with p-values $1.24\times 10^{-4}$ (Qwen-JVTON), $2.03\times 10^{-12}$ (OmniTry), $9.7\times 10^{-7}$ (Qwen-image-edit), $3.05\times 10^{-10}$ (InsertAnything), and $1.61\times 10^{-50}$ (Any2AnyTryOn). Significant tests further indicate JewelTry obtains more balanced performance on object consistency and background preservation, while also reach to a better scale-aware ability.

\begin{figure*}[t]
    \centering
    \includegraphics[width=\textwidth]{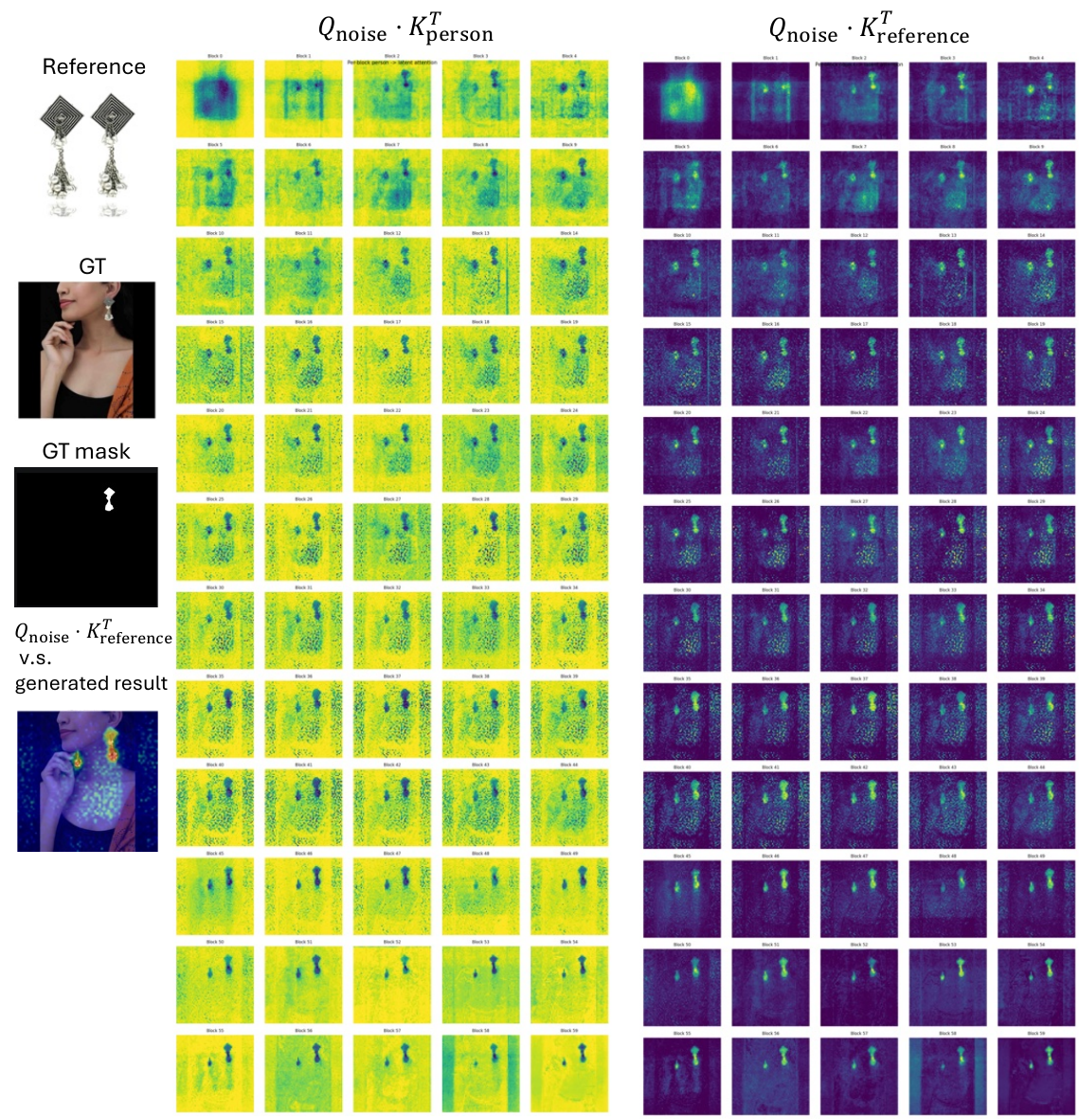}
    \caption{A visualization example for attention maps cross different block for earring. The left hand side visualizes $Q_{\text{noise}}\cdot K^T_{\text{person}}$ attention map cross the whole attention blocks averaged time steps during inference; While the right hand size visualizes $Q_{\text{noise}}\cdot K^T_{\text{reference}}$ attention map cross the whole attention blocks averaged time steps during inference.}
    \label{fig:attn_refine_vis}
\end{figure*}

\begin{figure*}[t]
    \centering
    \includegraphics[width=\textwidth]{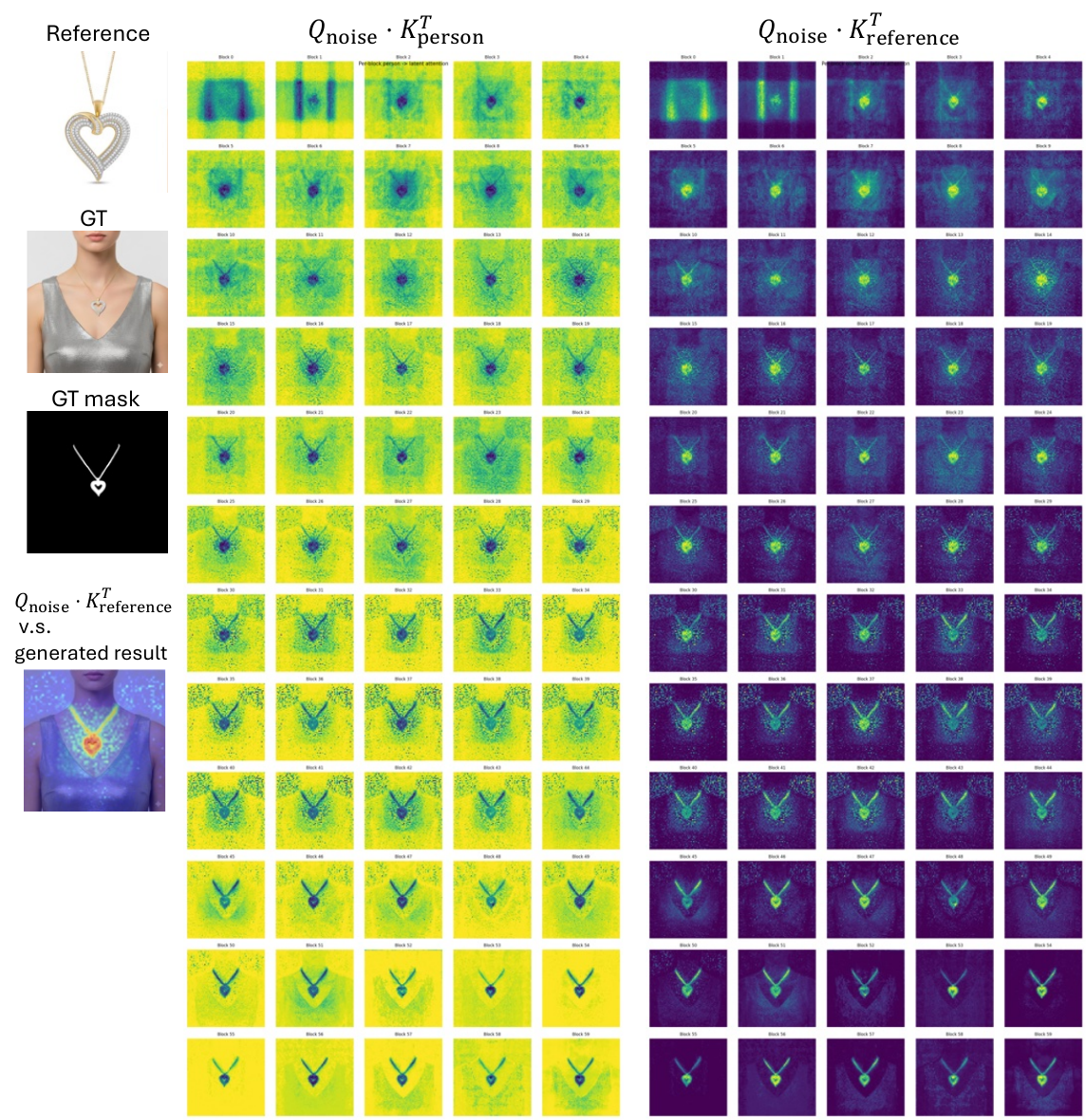}
    \caption{A visualization example for attention maps cross different block for necklace. The left hand side visualizes $Q_{\text{noise}}\cdot K^T_{\text{person}}$ attention map cross the whole attention blocks averaged time steps during inference; While the right hand size visualizes $Q_{\text{noise}}\cdot K^T_{\text{reference}}$ attention map cross the whole attention blocks averaged time steps during inference.}
    \label{fig:attn_refine_vis2}
\end{figure*}

\section{Attention Refinement Loss}\label{A:attn_loss}
In this section, we provide more technical details for the proposed attention refinement loss.

\subsection{Attention refinement loss and block-wise cross attention}
We provide further discussion and insights of the proposed attention refinement loss. Inspired by~\citet{shin2025exploring}, attention refinement loss is motivated by the empirical finding of the effectiveness of block-wise attention between conditions and latent in MMDiT transformer. Figure~\ref{fig:attn_refine_vis} and Figure~\ref{fig:attn_refine_vis2} visualize the attention maps across MMDiT transformer blocks for $Q_{\text{noise}}\cdot K^T_{\text{person}}$ and $Q_{\text{noise}}\cdot K^T_{\text{reference}}$, averaged over diffusion time steps. We observe that attention maps from later transformer blocks (e.g., the last two rows in $Q_{\text{noise}}\cdot K^T_{\text{reference}}$ figures) more clearly highlight the target jewelry region in the noisy latent space, with substantially reduced background noise. This observation motivates our attention refinement loss: by extracting attention maps from later transformer blocks and supervising them with the ground-truth jewelry mask, we explicitly encourage the model to focus on fine-grained jewelry regions during training, thereby improving object localization and structural consistency.

\subsection{Timestep binary gate}~\label{sec:b_gate}
In the attention refinement loss, we supervise the cross-attention maps with the ground-truth object mask via a combined BCE and Dice objective. Since spatial structure is unreliable at high noise levels, we restrict this supervision to low-noise timesteps through a binary gate $\mathbf{1}[\sigma_b < \tau]$, which activates the loss only when the sample's noise level $\sigma_b$ falls below a threshold $\tau$ (we use $\tau = 0.5$, i.e. the lower half of the noise schedule). Samples with $\sigma_b \ge \tau$ contribute zero attention loss. The timestep binary gate is defined as 
\begin{equation}
    \mathbf{1}[\sigma_b <\tau] =   \left \{
    \begin{array}{cc}
         1 & \sigma_b < \tau; \\
         0 &\sigma_b \ge \tau;
    \end{array}
    \right.
\end{equation}

\section{Person-conditioned Classifier-Free Guidance}

\begin{figure*}[t]
    \centering
    \includegraphics[width=\textwidth]{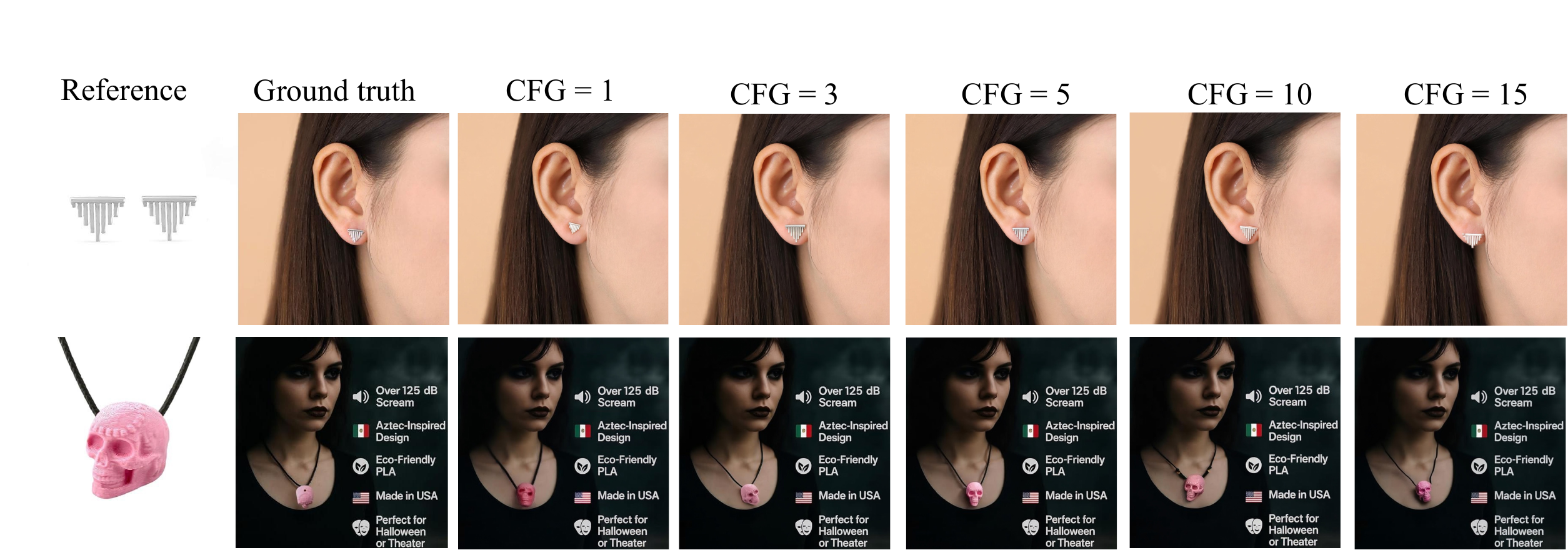}
    \caption{JewelTry inference result using different values of the classifier-free guidance scale.}
    \label{fig:CFG}
\end{figure*}

In the inference stage of JewelTry, we treat the scale token, text tokens, and reference jewelry tokens as a unified conditioning signal and apply classifier-free guidance (CFG) with a shared guidance value. Specifically, the guided noise prediction is formulated as:

\begin{multline}
\tilde{\epsilon}(z_t, c_s, c_{\text{txt}}, c_P, c_J)
=
\epsilon_\theta(z_t, c_s, c_{\text{txt}}, c_P, c_J)
+ \\
w \left[
\epsilon_\theta(z_t, c_s, c_{\text{txt}}, c_P, c_J)
-
\epsilon_\theta(z_t, c_P)
\right],
\end{multline}

where $c_s$, $c_{\text{txt}}$, $c_P$, and $c_J$ denote the scale, text, person image, and reference jewelry image conditions, respectively. $w$ is the guidance strength. In person-conditioned CFG, the person condition $c_P$ is retained in the negative sample, while the scale, text, and reference jewelry conditions are dropped, encouraging the model to follow the jewelry-related conditions while preserving the target person structure.

Figure~\ref{fig:CFG} presents a qualitative comparison of JewelTry inference results under different CFG values. The results show that a low CFG value, e.g., CFG =1, provides insufficient conditional guidance, making the generated try-on results less responsive to the specified product scale. In contrast, an overly large CFG value, e.g., CFG =10 or 15, can over-amplify the scale, text, and reference jewelry conditions, leading to unrealistic try-on results or structural collapse of the jewelry. These observations suggest that a moderate CFG value is necessary to balance scale controllability, jewelry fidelity, and visual realism.

\section{More study on single direction condition attention mechanism}

\begin{table}[ht]
  \centering
    \renewcommand{\arraystretch}{1}
    \renewcommand{\tabcolsep}{0.7mm}
{\fontsize{8pt}{9pt}\selectfont
  \begin{tabular}{l|l|ll|lll}
    \toprule
    Method & Mask type &DINO$_{\text{p}}$$\uparrow$ & LPIPS$_{\text{p}}$$\downarrow$ &  DINO$_{\text{ref}}\uparrow$  & DINO$_{\text{tar}}\uparrow$ & IoU$\uparrow$ \\
    \midrule
    Qwen-JVTON & - & 0.950 & 0.128 & 0.525 & 0.759  & 0.591\\
    \midrule
    Qwen-JVTON & $M$ &  0.953 & 0.121 & 0.559 & 0.760 & 0.589 \\
    JewelTry  & $M$ & 0.959 & 0.121  &  0.565 &  0.792 &  0.658 \\
    \midrule
    Qwen-JVTON & $M^{\text{P\&J}}$ & 0.952 & 0.151 & 0.555 & 0.756 & 0.504\\
    JewelTry  & $M^{\text{P\&J}}$ & 0.957 & 0.164  & 0.568 & 0.793 & 0.660\\
    \bottomrule
  \end{tabular}}
  \caption{Ablation study for single direction condition attention mask for block-wise attention. Where $M$ is the single direction condition attention mask defined in our main text.}
  \label{tab:SDAttn}
\end{table}

In JewelTry, we apply single-direction conditional attention to block the information flow from noisy latent tokens to jewelry condition tokens during training. Our ablation studies show that this mechanism improves coarse-level jewelry consistency during inference. In this section, we further investigate the effect of single-direction conditional attention under different block-wise masking strategies.

In the VTON task, the generated try-on result is expected to preserve both the identity and appearance of the person image while maintaining consistency with the reference jewelry image. Based on the hypothesis that information flow from noisy latent tokens to conditional tokens may cause condition collapse, we further examine whether blocking such information flow to both person and jewelry tokens can improve background preservation and object consistency. Following the notation in the main text, we define a mask $M^{\text{P\&J}}$ that blocks the reverse attention paths from noisy latent tokens to both person and reference jewelry tokens as follows:
\begin{equation}
    M_{ij}^{\text{P\&J}} = \left \{
    \begin{array}{cc}
         - \infty & (i \in \text{P},\; j \in \text{noise}) \cup (i \in \text{J},\; j \in \text{noise}); \\
         0 &  \text{otherwise};
    \end{array}
    \right.
\end{equation}

\begin{figure*}[t]
    \centering
    \includegraphics[width=0.8\textwidth]{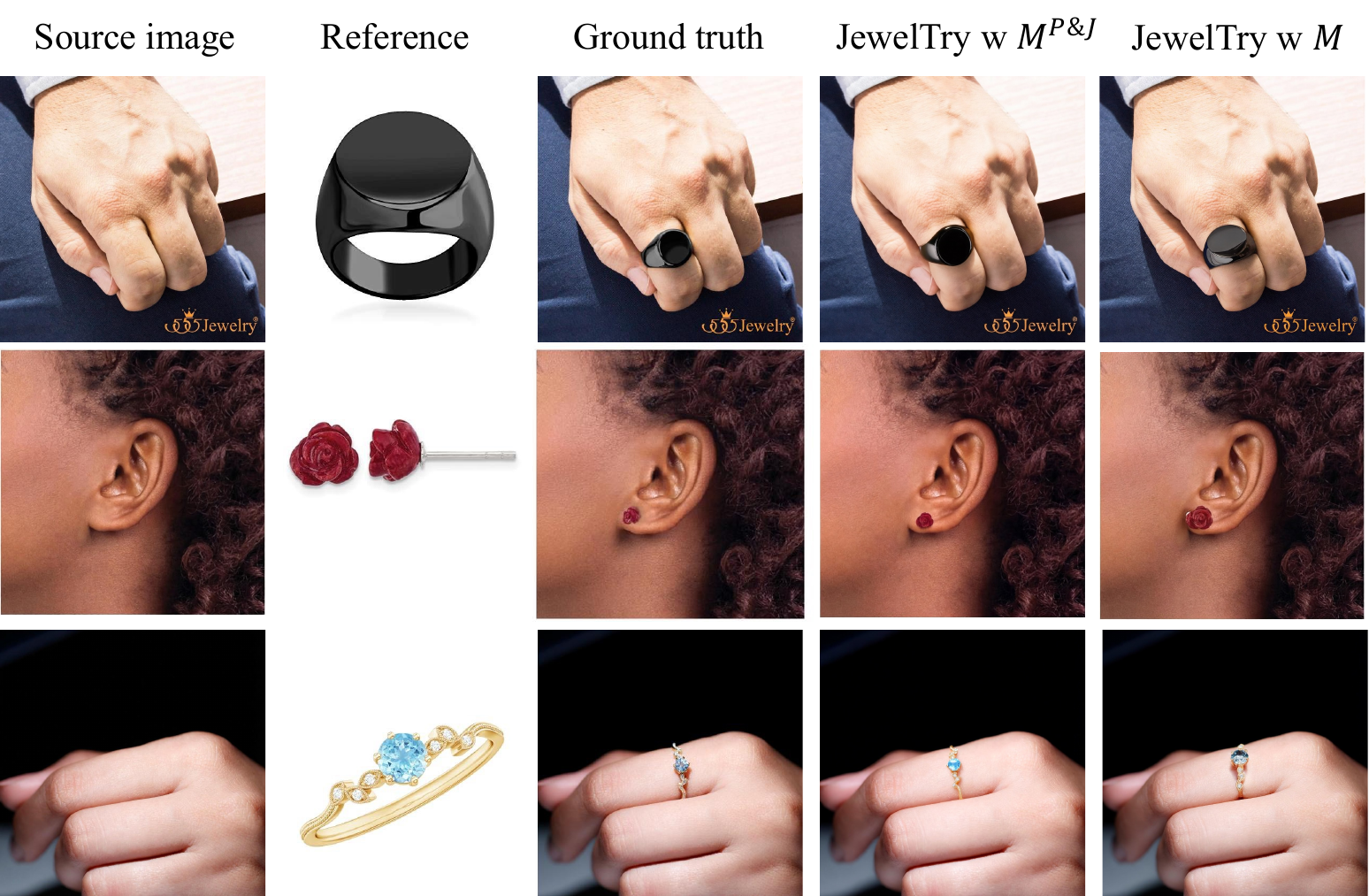}
    \caption{Qualitative comparison of different single direction condition attention mechanism on JewelTry. Although JewelTry with $M^{P\&J}$ has better object consistency; however, it suffers bad background preservation, skin tone color shift, and loss fine details on input source images.}
    \label{fig:SDAttn}
\end{figure*}

The mask $M^{\text{P\&J}}$ blocks the reverse attention paths from latent noisy tokens to person and reference jewelry tokens, with quantitative results reported in Table~\ref{tab:SDAttn}. In our ablation study, we integrate a single direction condition attention mask to both Qwen-JVTON and JewelTry. The comparison shows that blocking person and jewelry tokens further improves object consistency; however, it slightly degrades background preservation with higher LPIPS$_{\text{p}}$. Figure~\ref{fig:SDAttn} provides a visual comparison between the two masking strategies. Although $M^{\text{P\&J}}$ achieves better object consistency, it introduces a noticeable color-shift issue, where the skin tone of the source image is slightly changed. This may be because suppressing attention to person tokens weakens the model’s access to person-specific appearance cues, making it less effective in preserving local color and illumination consistency.

\section{More visualization}

\begin{figure*}[t]
    \centering
    \includegraphics[width=\textwidth]{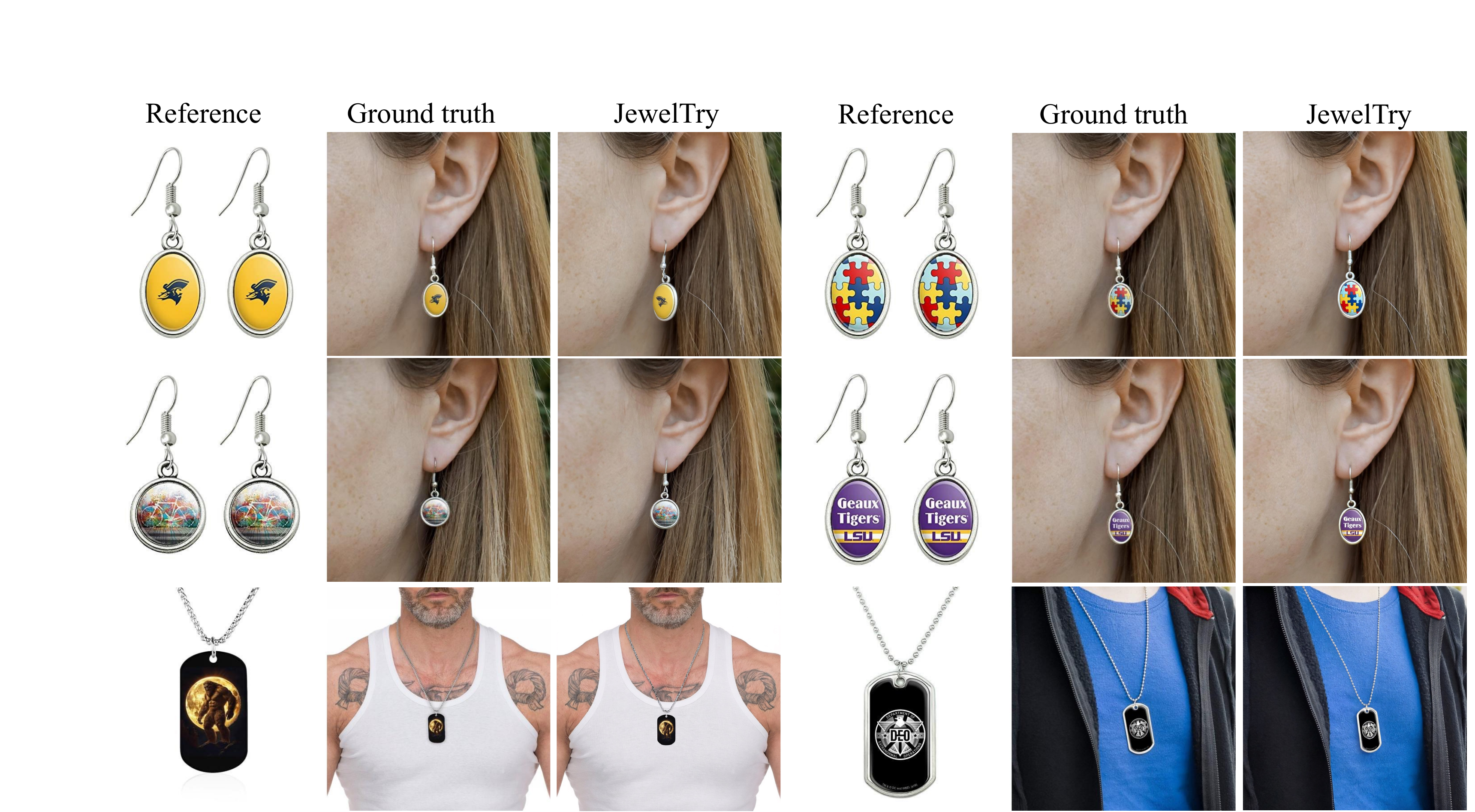}
    \caption{Additional JewelTry results on JVTO-Bench. These examples demonstrates JewelTry's ability to preserve fine-grained textual details in jewelry.}
    \label{fig:textual}
\end{figure*}

Many jewelry items, such as earrings and necklaces, require the model to preserve not only complex topology but also fine-grained textual details, similar to garments. As shown in Figure~\ref{fig:textual}, JewelTry effectively preserves these textual details in the generated jewelry.

\begin{figure*}[t]
    \centering
    \vskip -0.2in
    \includegraphics[width=\textwidth]{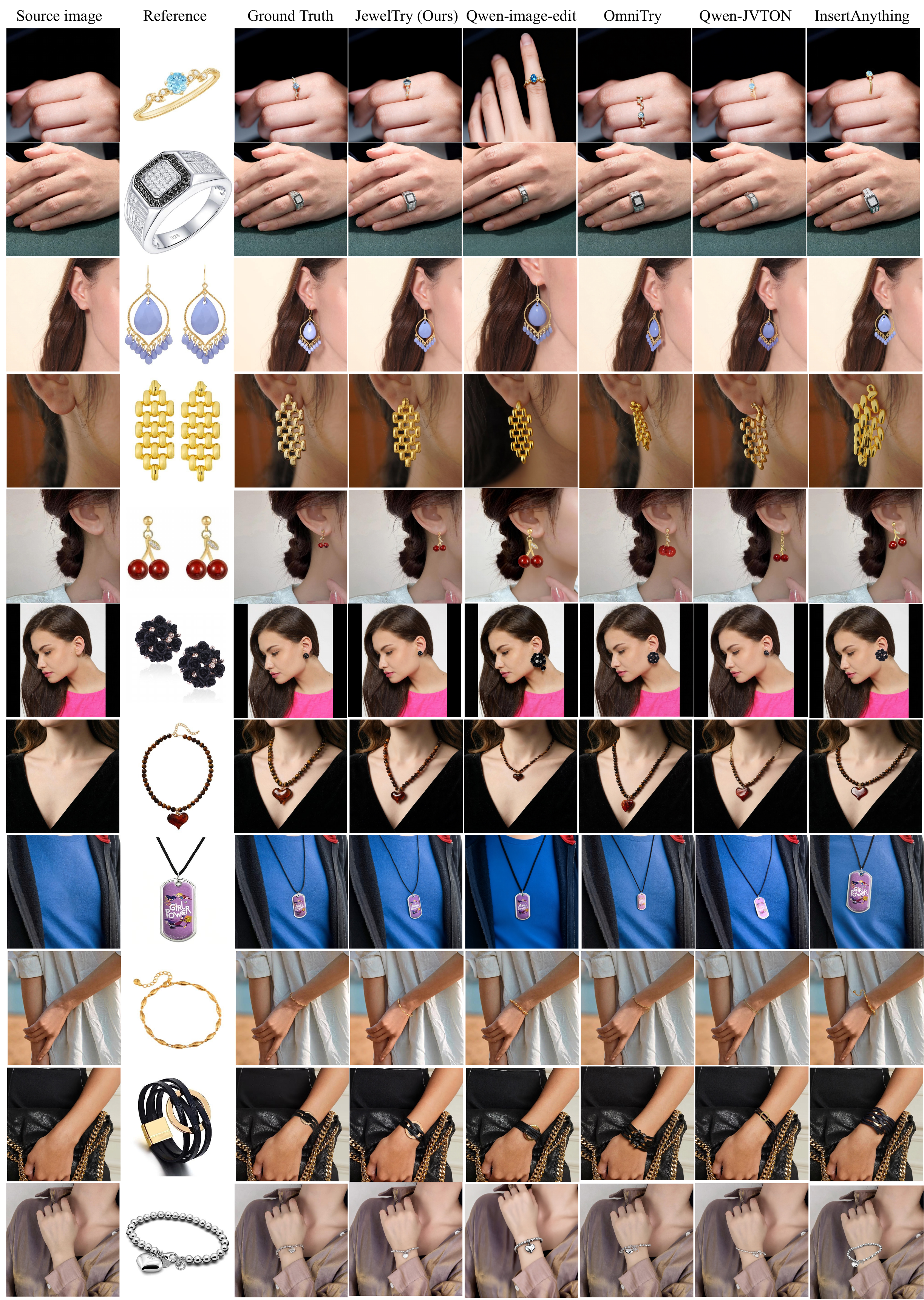}
    \caption{More visual comparison on JVTO-Bench cross four jewelry categories.}
    \label{fig:comp2}
\end{figure*}

\begin{figure*}[t]
    \centering
    \includegraphics[width=\textwidth]{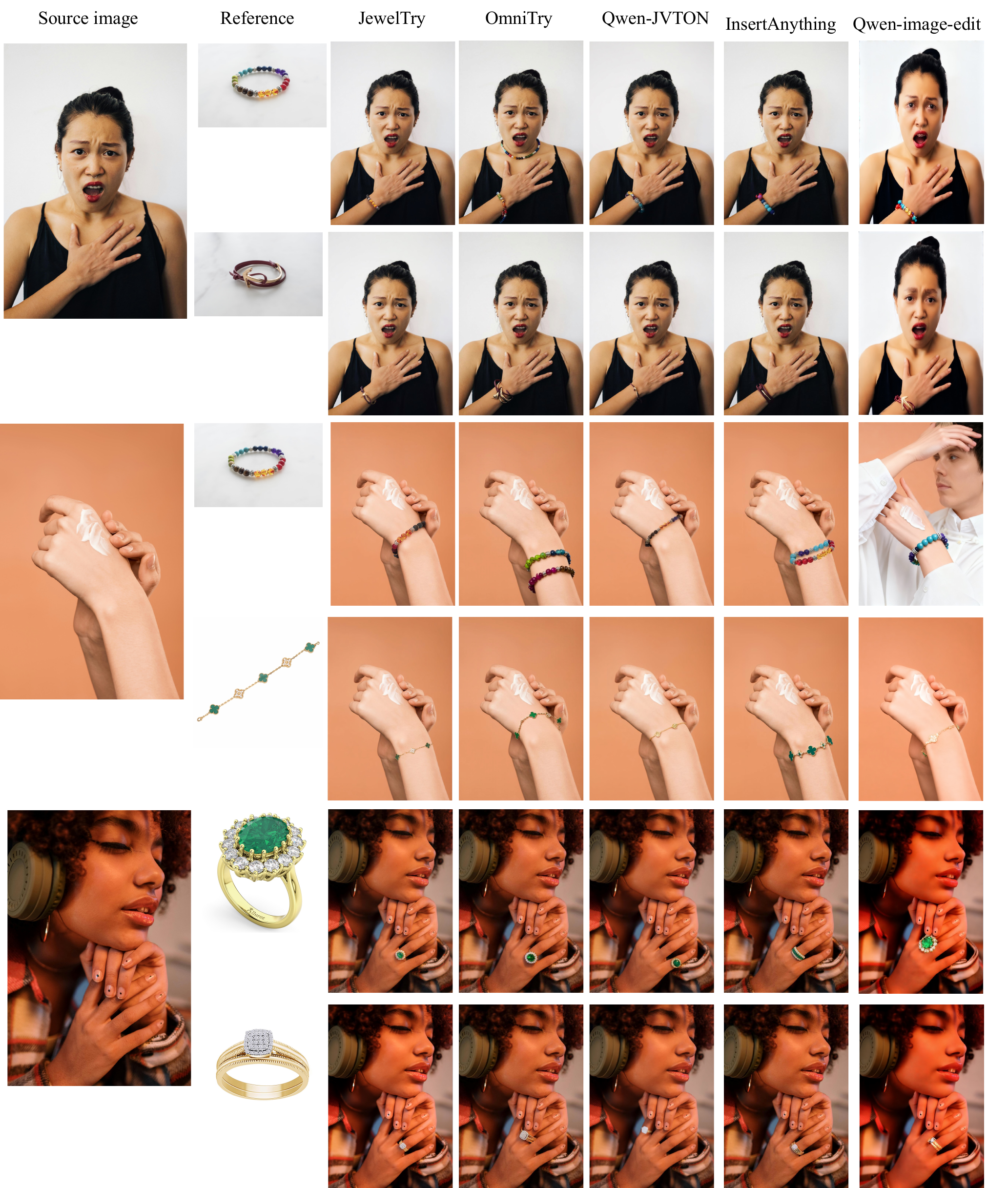}
    \caption{More visual comparison on OmniTry-Bench for bracelet and ring categories.}
    \label{fig:comp3}
\end{figure*}

\begin{figure*}[t]
    \centering
    \includegraphics[width=\textwidth]{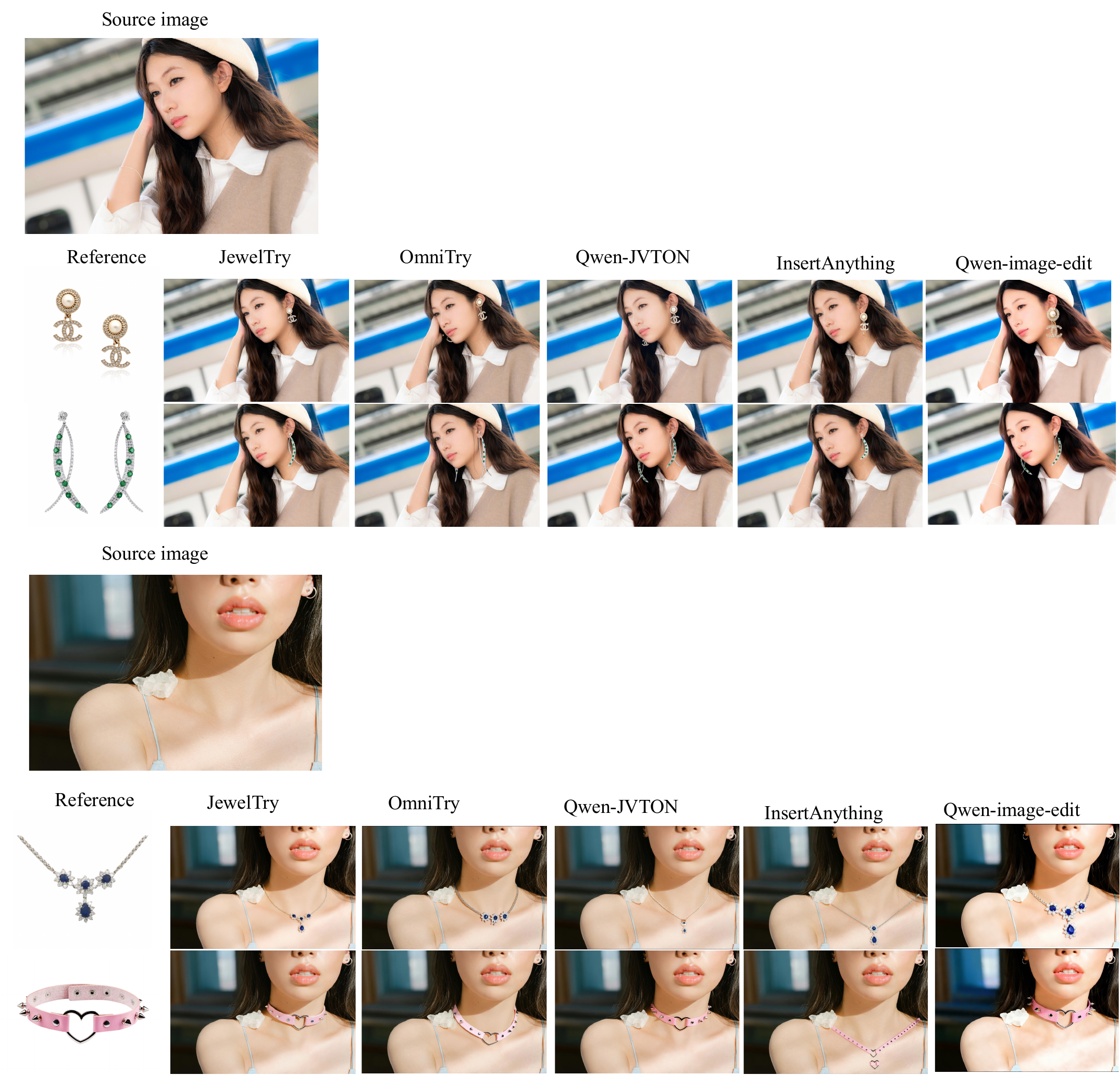}
    \caption{More visual comparison on OmniTry-Bench for earrings and necklace categories.}
    \label{fig:comp4}
\end{figure*}

We provide additional qualitative comparisons on both JVTO-Bench and OmniTry-Bench in Figure~\ref{fig:comp2} to Figure~\ref{fig:comp4}. For clarity, we compare JewelTry against the four most competitive baselines: OmniTry, Qwen-JVTON, Qwen-image-edit, and InsertAnything, because Any2AnyTryOn is not initially designed for jewelry VTON task. Although OmniTry-Bench does not provide product-scale annotations, JewelTry consistently generates jewelry with better structural fidelity, texture preservation, and fine-grained detail consistency. In particular, our method more faithfully preserves the geometry and appearance of the reference jewelry while maintaining natural integration with the wearer, demonstrating the effectiveness of the proposed object-consistency components.

\bibliography{aaai2027}
